%% file: acl_latex.tex
\pdfoutput=1

\documentclass[11pt]{article}

\usepackage[final]{acl}

\usepackage{times}
\usepackage{latexsym}

\usepackage[T1]{fontenc}

\usepackage[utf8]{inputenc}

\usepackage{microtype}

\usepackage{inconsolata}

\usepackage{graphicx}

\input{packages}
\input{commands}

\title{\method: Hierarchical Agentic Framework for Multi-Page Visual Document Understanding}

\author{Yiqiao Jin$^1$\thanks{Work done as an intern at J.P. Morgan AI Research.}, Rachneet Kaur$^2$, Zhen Zeng$^2$, Sumitra Ganesh$^2$, and Srijan Kumar$^1$ \\
  \textsuperscript{1}Georgia Institute of Technology, \textsuperscript{2} J.P. Morgan AI Research \\
  \texttt{\{yjin328,srijan\}@gatech.edu}, \texttt{zhen.zeng@jpmchase.com}, \\
  \texttt{\{rachneet.kaur,sumitra.ganesh\}@jpmorgan.com}
  \\
  \url{https://SlideAgent.github.io/}
}

\begin{document}
\maketitle

\input{src/abstract}

\input{src/intro}

\input{src/method}

\input{src/evaluation}

\input{src/related}
\input{src/conclusion}

\input{src/limitation}

\input{src/disclaimer}

\bibliography{cite}

\appendix

\input{src/appendix}

\end{document}

%% file: packages.tex
\usepackage{algorithm}
\usepackage{algorithmicx}
\usepackage{algpseudocode}
\usepackage{amsmath}
\usepackage{amsthm}
\usepackage{array}
\usepackage{booktabs}
\usepackage{calc}
\usepackage{colortbl}
\usepackage{enumitem}
\usepackage{footnote}
\usepackage{graphicx}
\usepackage{hyperref}
\usepackage{cleveref}
\usepackage{listings}
\usepackage{mathtools}
\usepackage{microtype}
\usepackage{multirow}
\usepackage{newfloat}
\usepackage{seqsplit}
\usepackage{subcaption}  
\usepackage{url}
\usepackage{xcolor}
\usepackage{xspace}
\usepackage{lineno}
\usepackage{pifont}
\usepackage{wrapfig}
\usepackage{cleveref}
\usepackage{lipsum}

%% file: commands.tex
\usepackage{algorithm}
\usepackage{algorithmicx}
\usepackage{algpseudocode}
\usepackage{amsmath}
\usepackage{amsthm}
\usepackage{array}
\usepackage{booktabs}
\usepackage{calc}
\usepackage{colortbl}
\usepackage{enumitem}
\usepackage{footnote}
\usepackage{graphicx}
\usepackage{hyperref}
\usepackage{listings}
\usepackage{mathtools}
\usepackage{microtype}
\usepackage{multirow}
\usepackage{newfloat}
\usepackage{seqsplit}
\usepackage{subcaption}  
\usepackage{url}
\usepackage{xcolor}
\usepackage{xspace}
\usepackage{lineno}
\usepackage{wrapfig}
\usepackage{lipsum}
\usepackage{tabularx}

\newcommand{\method}{SlideAgent\xspace}

\newcommand{\green}[1]{\textcolor[HTML]{377771}{#1}}

\definecolor{lightergray}{gray}{0.9}

%% file: src/abstract.tex
\begin{abstract}
Multi-page visual documents such as 
manuals, brochures, presentations, and posters convey key information through layout, colors, icons, and cross-slide references. 
While multimodal large language models (MLLMs) offer opportunities in document understanding, current systems struggle with complex, multi-page visual documents, particularly in fine-grained reasoning over elements and pages. 
We introduce \method, a versatile agentic framework for understanding \emph{multi-modal}, \emph{multi-page}, and \emph{multi-layout} documents, especially slide decks. 
\method employs specialized agents and decomposes reasoning into three specialized levels--global, page, and element--to construct a structured, \emph{query-agnostic} representation that captures both overarching themes and detailed visual or textual cues. During inference, \method selectively activates specialized agents for multi-level reasoning and integrates their outputs into coherent, context-aware answers.
Extensive experiments show that \method significantly improves accuracy over both proprietary (+7.9\%) and open-source models (+9.8\%). 
\end{abstract}

%% file: src/intro.tex
\section{Introduction}
Visual documents--from earnings reports and academic lectures to business strategy presentations--are ubiquitous, conveying ideas not only from text, but also from the intricate interplay of layout, icons, visual hierarchy, and cross-page relationships. 
These documents are central to high-stakes domains such as finance, science, and technology, and small misunderstandings can cascade into costly decisions: analysts extract key metrics from earnings decks,  engineers review tech specs to avoid implementation errors, and educators generate study materials whose inaccuracies can propagate at scale. 
Accurately interpreting them thus remains a pressing challenge.

\begin{figure}
\centering
\includegraphics[width=0.97\linewidth]{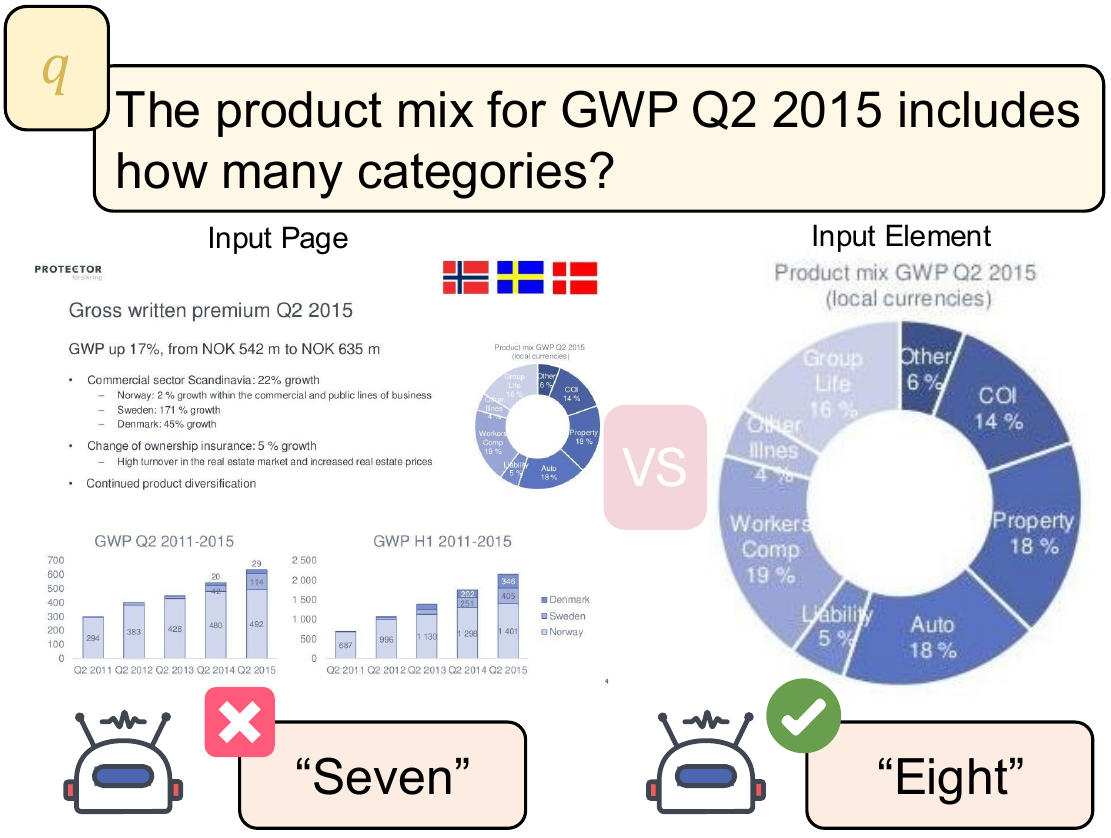}
\caption{
Accurate element parsing is important for fine-grained visual document understanding. 
The MLLM miscounts product-mix categories when reasoning over the entire slide, but succeeds after isolating the chart. 
}
\label{fig:teaser}
\vspace{-6mm}
\end{figure}

\paragraph{Challenges.} 
Accurately interpreting these \emph{multi-page}, \emph{multi-modal} artifacts remains challenging.
Recent advances in multimodal large language models (MLLMs) have accelerated document understanding~\citep{verma2024adaptagent}, yet three gaps remain:
\noindent \textbf{1) Scalable Fine-Grained Reasoning.} 
State-of-the-art MLLMs process a limited number of images at once~\citep{liu2023visual} and tend to treat each page holistically, missing fine-grained, element-level cues required for user queries~\citep{faysse2024colpali,tanaka2025vdocrag}. 
In Figure~\ref{fig:teaser}, an MLLM can miscount chart segments on a cluttered page of a slide deck, but correctly identifies all segments once the relevant chart is cropped--highlighting the importance of element-level parsing for latent reasoning abilities. 
\textbf{2) Domain-Specific Visual Semantics.} 
Most MLLMs are pre-trained on natural images~\citep{wu2024next}, lacking exposure to domain-specific diagrams, financial charts, or scientific plots. Consequently, they struggle with the specialized language of visual documents~\citep{cho2024m3docrag}.  
For example, logos appear on every page, reinforcing brand identities but do not offer additional content. Color schemes encode categorical information (red for losses and green for gains in financial reports). Icons convey abstract concepts (lightbulbs for innovation, arrows for causal relationships); and spatial positioning signals importance (centered elements typically matter more than corner annotations). 
However, current MLLMs 
falter in spatial reasoning~\cite{wang2024picture,wang2024docllm}, failing to locate visual elements~\cite{sharmathink,bhattacharyya2025information,polak2025leveraging}. 
Low-resolution visual encoders in MLLMs~\citep{liu2023visual} further miss details such as footnotes or superscripts. 
\textbf{3) Metadata-Free Integration.} Many systems
~\cite{Singer-Vine_pdfplumber_2025,huridocs2025pdf,rausch2021docparser}
rely on clean metadata--figure locations, hierarchy tags, embedded text layers--that can be unavailable or corrupted in real-world PDF. 
Users may take screenshots of PDF documents, scan copies of physical documents, export slides or documents as flattened PDFs, upload or share PDFs generated from software that strips out or does not preserve document structure. 
Recent metadata-free methods~\citep{yu2024visrag,tanaka2025vdocrag} address these by parsing only visual images without relying on the metadata, despite a performance gap. 



\vspace{-1mm}
\paragraph{This Work.} We present \method, an MLLM-based agentic framework for \textbf{fine-grained understanding of multi-page, varying-size visual documents}. 
Inspired by the human information processing model~\citep{lang2000limited, naysmith2021neural}, 
\method employs a hierarchical architecture with specialized agents at three levels: \emph{global} (document-wide topics), \emph{page} (page-specific features and cross-page relations), and \emph{element} (fine-grained components such as charts, figures, and text blocks). 
During \emph{knowledge construction} stage, \method parses layout and generates \emph{query-agnostic} knowledge at each level. 
At \emph{inference} stage, \method retrieves \emph{query-specific} knowledge, enabling scalable fine-grained reasoning over relevant pages and elements. 
We benchmark \method and baseline models, demonstrating significant performance on both open-source and proprietary models. \method surpasses its MLLM counterpart by 7.9\% (proprietary) and 9.8\% (open-source) in accuracy, respectively. 
We also show that \method enhances spatial reasoning, visual counting, and cross-element understanding, with results that are highly interpretable. 


%% file: src/method.tex
\vspace{-1mm}
\section{Method}
\vspace{-1mm}


\paragraph{Problem Formulation}
Given a multi-page visual document $\mathcal{P}= \{p_1, \ldots, p_{|\mathcal{P}|}\}$ with $|\mathcal{P}|$ pages and a query $q$, the goal is to generate a natural language answer $a = f(q, \mathcal{P})$ by reasoning over relevant visual and textual elements. 

\begin{figure*}[htbp]
\centering
\includegraphics[width=0.97\linewidth]{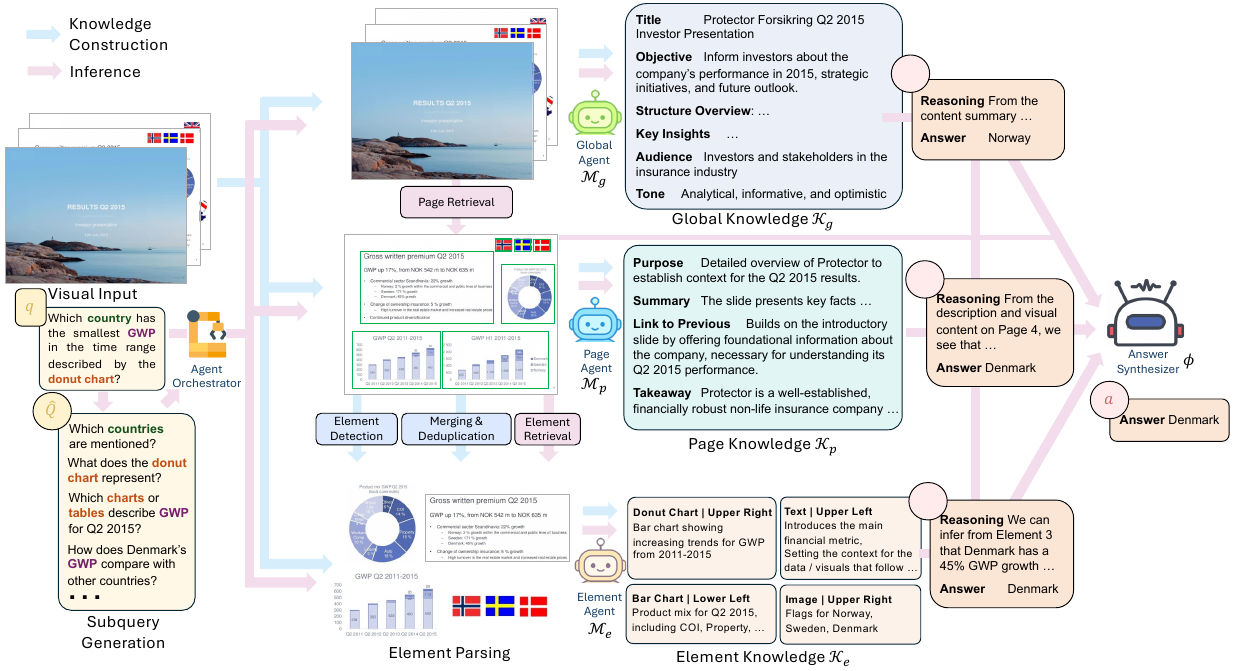}
\vspace{-2mm}
\caption{
\method generates knowledge about input slide decks in a hierarchical manner at 3 levels: \emph{global}, \emph{page}, and \emph{element}. At each level, specialized agents generate query-agnostic knowledge during \emph{knowledge construction}, then retrieve and reason over query-specific textual \& visual knowledge during \emph{inference} stage. Sample knowledge $\mathcal{K}$ generated by \method is in Appendix Figure~\ref{fig:sample_global_knowledge},\ref{fig:sample_page_knowledge},\ref{fig:sample_element_knowledge} and answers generated by the agents are in Figure~\ref{fig:sample_answer}. 
}
\label{fig:model}
\vspace{-5mm}
\end{figure*}

\vspace{-2mm}
\paragraph{Overview}
As shown in Figure~\ref{fig:model}, \method operates in two stages: 
1) \emph{Knowledge Construction}: Build a hierarchical, \emph{query-agnostic} knowledge base $\mathcal{K} = \{\mathcal{K}_g, \mathcal{K}_p, \mathcal{K}_e\}$ capturing global, page, and element knowledge;
2) \emph{Retrieval and Question-Answering}: Using multi-level retrieval to retrieve \emph{query-specific} content from $\mathcal{K}$ and synthesize the answer $a$, 
ensuring both broad contextual understanding and fine-grained reasoning.

\subsection{Knowledge Construction Stage}
\label{sec:knowledge}
Given a multi-page document, \method constructs hierarchical knowledge at three levels using specialized agents in a top-down manner. 

\paragraph{Global Agent} 
The global agent $\mathcal{M}_g$ generates initial document-level knowledge $\mathcal{K}_g^{(0)}$, capturing the overall summary, objectives, and narrative flow of the document. This layer establishes overarching themes to support high-level reasoning about the document's purpose. Since visual documents are often large and MLLMs have a limited capacity for processing visuals, \method samples the first three pages to generate $\mathcal{K}_g^{(0)}$ (Appendix Figure~\ref{fig:sample_global_knowledge}). 

\paragraph{Page Agent}
For each page $p_i \in \mathcal{P}$, the page agent generates page-level knowledge $\mathcal{K}_p$ in a sequential manner, conditioned on the page's visual content $v_i$, the initial global knowledge $\mathcal{K}_g^{(0)}$, and the knowledge from the preceding page $\mathcal{K}_p^{i-1}$: 
\begin{equation}
\mathcal{K}_p^{i} = \mathcal{M}_p (v_i, \mathcal{K}_g^{(0)}, \mathcal{K}_p^{i-1}), i \in [1, |\mathcal{P}|]. \mathcal{K}_p^{0} = \emptyset
\label{eq:page_agent}
\end{equation}
The complete page-level knowledge $\mathcal{K}_p = \bigcup_{i=1}^{|\mathcal{P}|} \mathcal{K}_p^{i}$ provides an intermediate representation that captures page-specific content while linking them to the global context (sample in Figure~\ref{fig:sample_page_knowledge}). 
To ensure comprehensive understanding of all pages, $\mathcal{K}_p$ is subsequently used to refine $\mathcal{K}_g^{(0)}$ through a single-pass fieldwise rewrite:
\begin{equation}
\mathcal{K}_g = \mathcal{M}_g^{\mathrm{refine}}(\mathcal{K}_g^{(0)}, \mathrm{concat}(\mathcal{K}_p^1,\ldots,\mathcal{K}_p^{|\mathcal{P}|})).
\label{eq:global_refine}
\end{equation}
The global agent regenerates each global field from the complete page-level evidence, using $\mathcal{K}_g^{(0)}$ only as a weak prior for terminology and early context. This rewrite can overwrite earlier hypotheses when later pages provide stronger evidence, reducing bias toward the first pages. The refinement prompt is shown in Appendix Table~\ref{tab:global_refine_prompt}. 
\paragraph{Element Agent} 
MLLMs often struggle with spatial reasoning over visual documents~\cite{wang2024picture,wang2024docllm}, failing to locate visual elements~\cite{sharmathink,bhattacharyya2025information,polak2025leveraging}. To address this, our element agent integrates external tools to explicitly capture the spatial and structural information of each page.

\input{tables/tab-query-cases}

At the finest granularity, the element agent decomposes each page $p_i$ into a set of elements using a layout parsing pipeline $f: v_i \rightarrow \{(i, e_j, b_j, t_j)\}_{j=1}^{M_i}$, where each element is represented by its page index $i$, verbatim text $e_j$, bounding box coordinates $b_j$, and element type $t_j$. 
$f$ integrates text detection, layout detection, and element classification, followed by post-processing to merge fragmented elements (Appendix~\ref{app:merge}). 

For each detected element $e_j$, the agent consumes the annotated visuals and metadata to generate element-level knowledge: 
\begin{equation}
    \mathcal{K}_e^{j} = \mathcal{M}_e(i, v_i, e_j, b_j, t_j, \mathcal{K}_g, \mathcal{K}_p^{i}).
    \label{eq:element_agent}
\end{equation}
$\mathcal{K}_e^{j}$ consists of the element's semantic role, functional purpose, and its relation to the slide page (example in Figure~\ref{fig:sample_element_knowledge}). 
This design allows \method to reason consistently across diverse visual content while preserving spatial relationships for document understanding.

\vspace{-1mm}
\subsection{Inference Stage}
\label{sec:inference}
\vspace{-1mm}
\paragraph{Query Classification} 
Different queries require different perspectives to answer effectively. For instance, queries about global understanding, like ``What is the overall theme of this presentation?'' requires a broad overview from a macroscopic perspective and activates only the global agent. In contrast, a fact-based query, like ``What is the revenue on slide 3?'' requires detailed, slide-specific information and requires both the page and element agents. Leveraging too many agents may increase computation or introduce noise. Thus, the agent orchestrator first attempts to classify each query $q$ into one of 4 predefined categories, such as global understanding, fact-based direct queries, multi-hop reasoning, and layout \& visual relationships. Each type corresponds to a question-answering strategy and a targeted set of agents (Table~\ref{tab:query_cases}). 
For instance, global understanding queries activate only the global agent, as they require a broad overview, while fact-based queries trigger both the page and element agents to retrieve detailed, slide-specific information. 
If none of the predefined categories apply, the agent defaults to the ``unknown'' category, which activates all agents. 


\paragraph{Subquery Generation and retrieval} 
The original query $q$ is usually short and can lead to noisy retrieval. Using $q$, \method generates subqueries $\hat{Q}$ targeting key entities in the query. For example, for the query `Which \emph{country} has the smallest \emph{GWP} in the time range described by the \emph{donut chart},' the model generates subqueries related to keywords such as \emph{country}, \emph{donut chart}, and \emph{GWP}. 
$q$ and $\hat{Q}$ are concatenated 
to jointly retrieve the top-$k_\ell$ pages $\hat{\mathcal{P}}$ and the top elements $\hat{\mathcal{E}}$ along with their page / element knowledge $\mathcal{K}_p$ and $\mathcal{K}_e$. Note that the retriever can include simple sparse retrievers such as BM25~\cite{robertson2004simple}, dense retriever such as SFR~\cite{SFRAIResearch2024}, GTE~\citep{li2023towards} and Linq~\citep{LinqAIResearch2024}, and multimodal retriever such as COLPALI~\cite{faysse2024colpali} and VisRAG~\citep{yu2024visrag}. 




\paragraph{Answer Generation and Synthesis} 
The system guides structured reasoning through hierarchical context understanding to generate $h_g, h_p, h_e$, which contain both the agent's answer and its reasoning. 
The global context is processed to generate  $h_g = f_g(\mathcal{K}_g, \mathcal{P}_s, q)$, which captures the overall document-level context. 
The page-level agent then derives from the retrieved pages and corresponding knowledge: 
$h_p = f_p(\mathcal{K}_p, \mathcal{R}_p(\mathcal{P}, {\{q\} \cup \hat{Q}}), h_g)$. 
At the element-level, the visual input is annotated with bounding boxes $\{b_i\}$, which are then processed by the element agent to generate $h_e = f_e(\mathcal{R}'_e(\mathcal{E}, \{q\} \cup \hat{Q}), \mathrm{Annot}(\hat{V}))$, capturing the detailed visual and textual cues. 

If all agents agree according to answer matching (Appendix~\ref{app:answer_matching}), or only one agent is activated, the answer is taken directly from the activated agent. Otherwise, the answer synthesizer $\phi(\cdot)$ combines the reasoning from all agents and visuals from the retrieved pages to generate the final answer: 
\vspace{-1mm}
\begin{equation}
a = \phi(h_g, h_p, h_e, \{v_i : p_i \in \mathcal{R}(\hat{\mathcal{P}})\}). 
\end{equation}

%% file: tables/tab-query-cases.tex
\begin{table*}[htbp]
\centering
\small
\begin{tabular}{p{6cm}|p{6cm}|p{2.5cm}}
\toprule
\textbf{Case} & \textbf{Example} & \textbf{Agents} \\
\midrule
\textbf{Global Understanding} Asks about the overall theme, purpose, or general summary of the entire presentation. & "What is the main topic of the presentation?", "What is this deck about?" & Global \\

\textbf{Fact-based Direct Query}. Asks for specific facts, data, or information from particular slides. & "What is the revenue reported on slide 7?", "Which slide shows the product roadmap?" & Page, Element \\

\textbf{Multi-hop Reasoning} Requires comparing information across multiple slides or elements. & "Compare revenues in slide 5 and slide 10", "How do Sweden and Denmark compare?" & Global, Page, Element \\

\textbf{Layout / Visual Relationship} Asks about visual relationships, positioning, or layout elements. & "What does the diagram below the table on slide 12 illustrate?", "Is the color red used to denote negative performance?" & Element \\

\textbf{Uncertain} If the query is unclear, use all agents to answer the query. & \textbackslash  & Global, Page, Element \\
\bottomrule
\end{tabular}
\caption{\method's query classification that determines which hierarchical agents to activate.}
\label{tab:query_cases}
\vspace{-3mm}
\end{table*}

%% file: src/evaluation.tex
\section{Experiments}
\vspace{-1mm}

\input{tables/tab-overall_gpt4o}

\paragraph{Datasets} 
We evaluate \method and baselines on two tasks: 1) \emph{multi-page understanding}, using datasets such as SlideVQA~\citep{tanaka2023slidevqa}, TechSlides, and FinSlides~\citep{wasserman2025real}; and 2) single-page understanding, using InfoVQA~\citep{mathew2022infographicvqa}. Details of the datasets are in Appendix~\ref{app:dataset}. 


\paragraph{Models.} 
We benchmark \method against 3 types of baselines: 
1) \textbf{Multimodal LLMs}: 15 MLLMs from 8 model families, including proprietary models (GPT-4o~\citep{gpt4o}, Gemini~\citep{team2023gemini}, Claude~\citep{claude}) and open-source models (Llama-3.2~\citep{grattafiori2024llama}, InternVL3~\citep{chen2024internvl}, Phi-3~\citep{abdin2024phi}, Qwen2.5-VL~\citep{bai2025qwen2}); 
2) \textbf{Multimodal RAG Methods}: VisRAG~\cite{yu2024visrag}, VDocRAG~\cite{tanaka2025vdocrag}, and COLPALI~\cite{faysse2024colpali}; 3) \textbf{Multi-agent Systems}:  ViDoRAG~\cite{wang2025vidorag} and MDocAgent~\citep{han2025mdocagent}. 
We evaluate \method using two backbone MLLMs, including both proprietary models (GPT-4o~\citep{gpt4o}) and open-source models (InternVL3-8B~\citep{chen2024internvl}). These models are chosen for their widespread use in state-of-the-art QA systems~\cite{yu2024visrag,jin2025sara,cho2024m3docrag}. The parameter sizes, knowledge cutoff dates, and release dates are in Appendix Table~\ref{tab:models}. 
We use the text-based retriever SFR~\citep{SFRAIResearch2024} due to its strong efficiency-performance~\cite{cheng2024xrag,jin2025sara}. 
For models restricted to single-image input (e.g. LLaVA-v1.5~\citep{liu2023visual}), we concatenate the top 3 retrieved images into one following previous work~\cite{yu2024visrag}. 
In settings where ground-truth pages are available, we provide them as input to all models.  
Otherwise, each model receives as many retrieved pages up to its input capacity. 

\paragraph{Metrics.} 
We evaluate the performance of \method in end-to-end question answering. 
For questions asking about \emph{numeric} values, we extract, standardize, and compare the prediction and the ground-truth in various formats, including percentages, decimals, integers, and word-based representations (e.g., ``three'', ``thousand'', ``million''). 
Numbers are normalized to a unified format (e.g., `17k' $\rightarrow$ `17000', `2.5 million' $\rightarrow$ `2500000', `97\%' $\rightarrow$ `0.97). 
Otherwise, we use F1-score to evaluate the \emph{lexical overlap} between predicted and ground-truth answers. 
Both answers are normalized, tokenized, and preprocessed by removing stopwords and punctuation before metric calculation. 
For ranking, we use MRR, Hit@k, and nDCG@k (Appendix~\ref{app:metrics}) 

\vspace{-2mm}
\paragraph{Settings}
We evaluate \method under two realistic settings: 1) \textbf{End-to-End Performance.} The model must first retrieve relevant pages before answering the query, reflecting real workflows where users query long visual documents 
--analysts navigating 100-page earnings presentation, students reviewing lecture slides, or engineers inspecting multi-page technical specifications. 
This setting measures the end-to-end capability in retrieval, spatial reasoning, and layout understanding.
2) \textbf{Performance with Ground-truth Pages.} The model is directly given the page(s) containing the answer, isolating reasoning from retrieval. 
This mirrors cases where context is known--e.g., a product manager reviewing a linked design slide--and primarily focuses on reasoning and layout comprehension. 

\input{tables/tab-infovqa_proprietary}

\vspace{-1mm}
\subsection{End-to-End Performance}
\vspace{-1mm}
Tables~\ref{tab:performance_proprietary}/\ref{tab:infovqa_proprietary}/\ref{tab:performance_open_source} and Appendix Table~\ref{tab:infovqa_open_source} demonstrate the end-to-end performance of \method across proprietary and open-source models. 

\input{tables/tab-overall_internvl}

\paragraph{Consistent Improvements across Architectures} For both proprietary and open-source models, \method consistently outperforms all baseline methods (Type 2/3) and the base models 
across all metrics. 
For proprietary models (Table~\ref{tab:performance_proprietary}), \method improves overall accuracy by $+7.9$\%, numeric reasoning by $+8.3$\%, and lexical overlap by $+6.5$\% on SlideVQA. 
Similar trends are observed for TechSlides ($+7.5$\% overall; $+12.3$\% F1) and FinSlides ($+5.5$\% overall; $+17.5$\% F1), indicating robust performance across diverse domains. 
For open-source models, the advantage remains pronounced: \method outperforms the original MLLM by $+9.8$\% overall ($+11.7$\% numeric), showing that our hierarchical, multi-agent design generalizes well across MLLMs. 

\begin{figure}
\centering
\includegraphics[width=0.97\linewidth]{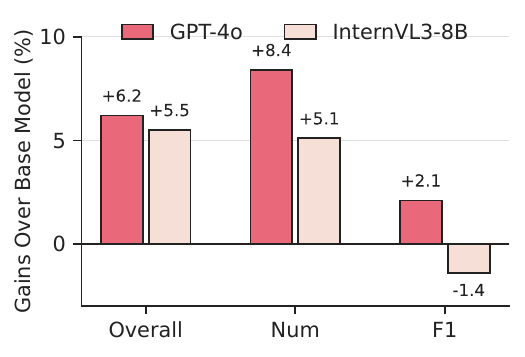}
\vspace{-3mm}
\caption{Absolute performance gains over MDocAgent~\cite{han2025mdocagent} when using GPT-4o and InternVL3-8B as the backbone models on SlideVQA~\cite{tanaka2023slidevqa}. \method yields significant overall improvements across both backbones.}
\label{fig:mdocagent_gains}
\vspace{-3mm}
\end{figure}

\paragraph{Comparison with Multimodal LLMs}
\method consistently achieves the best or second-best performance among proprietary models (Table~\ref{tab:performance_proprietary} and Appendix Table~\ref{tab:gt_pages_proprietary}). Notably, although the base model slightly lags behind stronger MLLMs such as Gemini-2.5-Flash in raw capability (e.g. 83.8 for Gemini-2.5-Flash vs. $77.0$\% for GPT-4o in Table~\ref{tab:performance_proprietary}), 
\method's structured reasoning pipeline fills this gap ($84.9$\% overall for SlideAgent). 
For open-source models, \method achieves better performance across all models except for the Qwen2.5-VL family. 
While strong base models such as Gemini-2.5 and Qwen2.5-VL already exhibit advanced multimodal comprehension, \method is model-agnostic in nature and can be directly applied to these models to further enhance performance.




\begin{figure}
\centering
\includegraphics[width=0.98\linewidth]{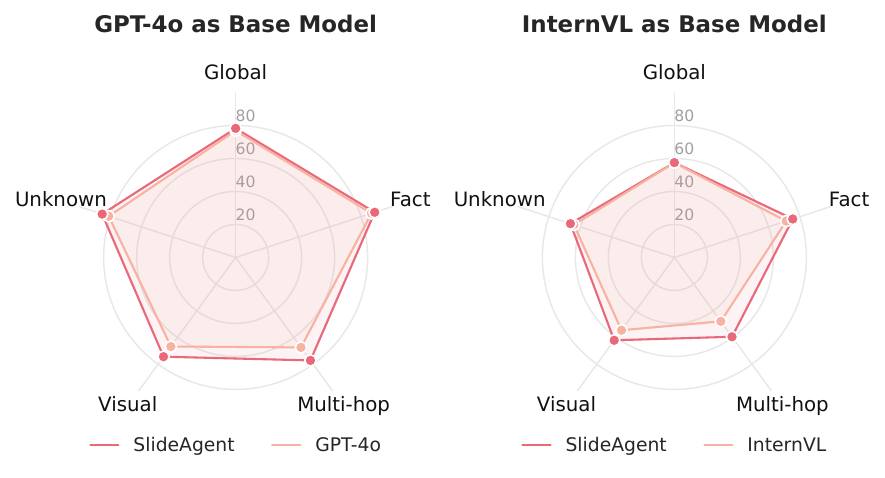}
\vspace{-1mm}
\caption{Accuracy of \method and base models (GPT-4o / InternVL3-8B) on different query types.}
\label{fig:correctness_by_case}
\vspace{-3mm}
\end{figure}

\vspace{-3mm}
\paragraph{Performance across Query Types}
Figure~\ref{fig:correctness_by_case} presents the performance breakdown across query types, as defined in Table~\ref{tab:query_cases}. 
\method improves question-answering across diverse query types, especially in multi-hop reasoning and visual/layout questions. 
The greatest improvement occurs on case 3 (multi-hop reasoning), with a $9.8$\% improvement ($67.4$\% to $77.2$\%). This means explicitly guiding the model's reasoning using generated knowledge $(\mathcal{K}_g, \mathcal{K}_p, \mathcal{K}_e)$ significantly improves reasoning capabilities. A notable $7.7$\% improvement ($66.7$\% $\rightarrow 74.4$\%) is also observed in visual/layout reasoning, demonstrating the benefit of fine-grained element-level reasoning and retrieval that multimodal LLMs hardly achieve.  
Both \method and its counterpart perform well on Case 2 (Fact-based Direct Queries), with only a modest $2.1$\% improvement, reflecting the model's ability to handle page-level reasoning with little space for further enhancement.

\subsection{Performance with Ground-truth Pages}
When ground-truth pages are provided (Table~\ref{tab:gt_pages_proprietary}/\ref{tab:gt_pages_internvl}), the performance gap narrows. As all models receive the exact pages with the answers, noise introduced in retrieval is eliminated. 
Under this oracle setting, \method still improves over the base model ($+7.7$\% overall and $+12.5$\% numeric on SlideVQA), demonstrating the effectiveness of element-level retrieval.

\subsection{Error Analysis}
\label{app:error_analysis}
\begin{figure}[htbp]
\centering
\includegraphics[width=0.97\linewidth]{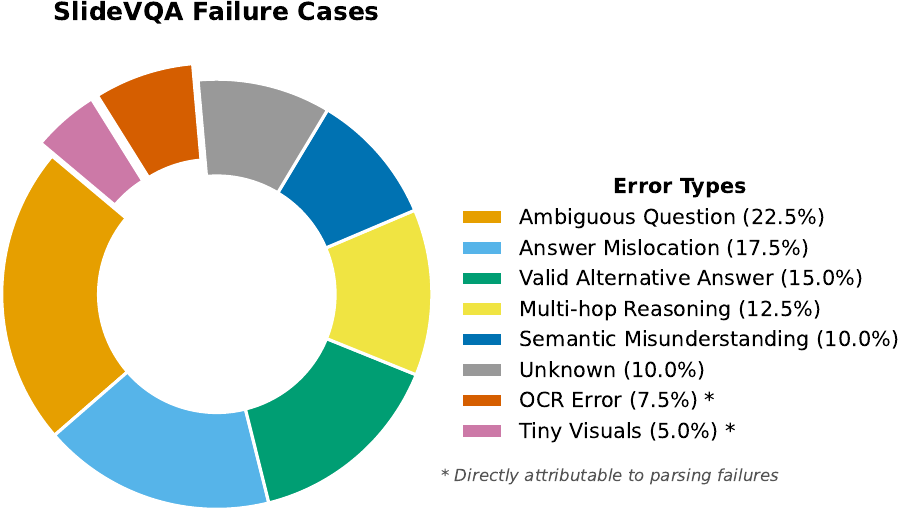}
\caption{
Error analysis of 40 failure cases on SlideVQA. Only OCR Error and Tiny Visuals (12.5\% combined) are directly attributable to parsing failures. 
}
\label{fig:error_cases}
\vspace{-3mm}
\end{figure}

We annotate 40 SlideVQA failure cases to quantify parsing-related error propagation. As shown in Table~\ref{fig:error_cases}, only OCR Error and Tiny Visuals are directly attributable to parsing failures, accounting for 5/40 cases (12.5\%). The dominant categories are Ambiguous Question (22.5\%), Answer Mislocation (17.5\%), and Valid Alternative Answer (15.0\%), indicating that ambiguity, retrieval drift, and annotation subjectivity account for a larger share of failures than OCR/layout parsing.

\subsection{Effectiveness of Knowledge Construction} 
\begin{figure}
\centering
\includegraphics[width=0.98\linewidth]{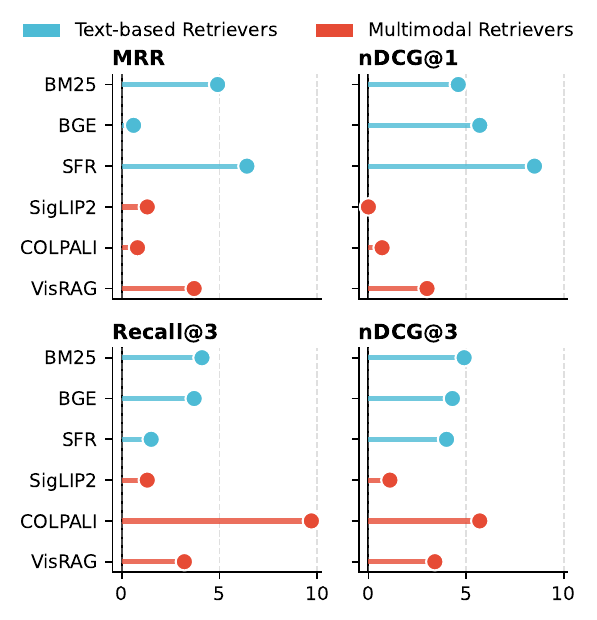}
\vspace{-3mm}
\caption{Artifacts generated by \method, particularly $\hat{Q}$ and $\mathcal{M}_p$, improves \emph{page-level} retrieval performance, especially for text-based retrievers. }
\label{fig:lollipop_retriever}
\vspace{-3mm}
\end{figure}

We evaluate whether hierarchical knowledge representations ($\mathcal{K}$) improve retrieval beyond end-to-end QA. Specifically, we test page-level retrieval using generated subqueries $\hat{Q}$ and page knowledge $\mathcal{K}_p$ from \method. Figure~\ref{fig:lollipop_retriever} shows significant gains across both text-based retrievers (BM25~\cite{robertson2004simple}, BGE~\cite{bge_embedding}, SFR~\cite{SFRAIResearch2024}) and multimodal retrievers (COLPALI~\cite{faysse2024colpali}, VisRAG~\cite{yu2024visrag}, SigLIP2~\cite{tschannen2025siglip}).

\vspace{-1mm}
\paragraph{Text-based Retrievers Show Largest Gains.} 
Structured agent outputs substantially enhance text-based retrievers. SFR achieves the largest gains (+6.4 MRR, +8.5 nDCG@1), showing that page-level knowledge $\mathcal{K}_p$ provides richer semantic signals than raw OCR, especially useful in the absence of the vision modality. 
Even sparse retrievers like BM25 improves (+4.9 MRR), as lexical matching benefits from structured page descriptions, which integrates multimodal cues that pure text extraction often miss. Notably, despite much lower computational costs, text-based retrievers already rival multimodal LLM-based methods, justifying our choice of SFR as the base retriever. 

\vspace{-1mm}
\paragraph{Multimodal Retrievers Exhibit Smaller but Consistent Gains.} COLPALI~\cite{faysse2024colpali} improves slightly in ranking quality (+0.8 MRR) but substantially in coverage (+9.7 Recall@3), indicating that structured subqueries help it surface more relevant pages even if they are not ranked at the very top. 
VisRAG~\cite{yu2024visrag} achieves a larger ranking boost ($+3.7$\% MRR), suggesting that multimodal LLMs still benefit from textual guidance in aligning queries to page content. 
SigLIP2 shows minimal gains ($+1.3$\% MRR), likely because it is optimized for natural image domains and transfers less effectively to document-style inputs.

\begin{figure}
\centering
\includegraphics[width=0.97\linewidth]{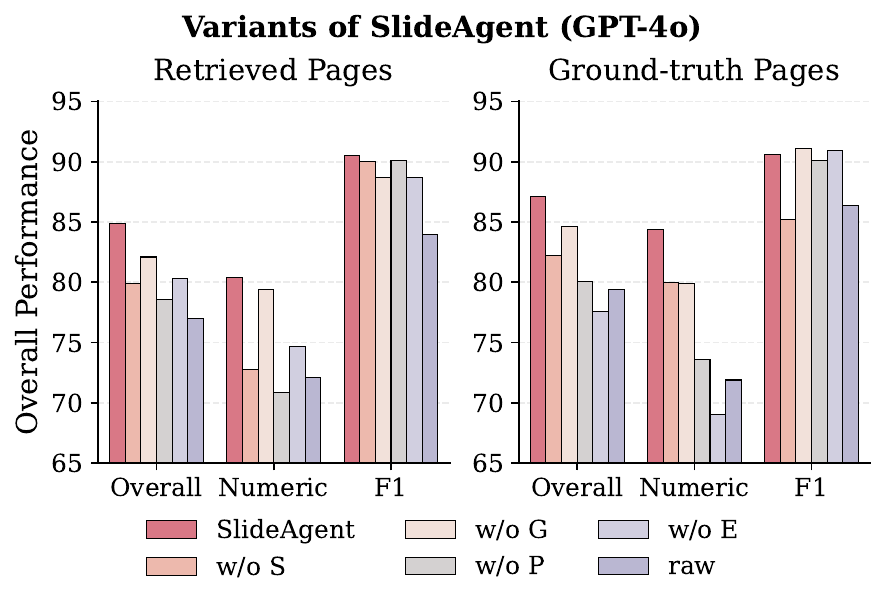}
\vspace{-2mm}
\caption{Performance comparison among variants of \method with base model GPT-4o. }
\label{fig:ablation}
\vspace{-4mm}
\end{figure}

\subsection{Ablation Studies}
\vspace{-1mm}
We analyze each design choice by removing global agent (w/o G), page agent (w/o P), element agent (w/o E), and subquery generation (w/o S). 
Figures~\ref{fig:ablation} summarize results across proprietary (GPT-4o) and open-source (InternVL3-8B) backbones. 

\paragraph{Page-level Reasoning is Critical} 
Removing the page agent causes the steepest degradation--GPT-4o drops –6.3 overall (–9.5 numeric) and InternVL3-8B –8.8 overall. The page agent integrates global themes $\mathcal{K}_g$ and sequential context $\mathcal{K}_p^{i-1}$ (Eq.~\ref{eq:page_agent}), enabling cross-slide coherence and multi-hop reasoning. Its absence breaks this structural bridge between global context and page details. 


\paragraph{Element-level Reasoning Ensures Precision}
Ablating the element agent causes a moderate yet consistent decline (–4.6 overall for GPT-4o; –6.3 for InternVL3-8B), especially numeric questions. Even with perfect retrieval, omitting fine-grained texts and layouts weakens factual grounding.

\paragraph{Impact of Global Thematic Guidance} 
Removing the global agent yields the smallest drop (-$2.8$\% GPT-4o; –$3.7$ InternVL3-8B), as lower-level agents already embed partial global context via $\mathcal{K}_p$ and $\mathcal{K}_e$ in knowledge construction (Eq.~\ref{eq:page_agent}/\ref{eq:element_agent}). However, responses become less aware of overarching themes. 

\paragraph{Subquery Generation Strengthens Retrieval.}
Removing subqueries causes larger losses under retrieval (–5.0 GPT-4o; –11.3 InternVL3-8B) than with ground-truth pages (–2.9 to –4.9). Visual-aware subqueries markedly improve retriever accuracy by aligning textual intent with visual semantics--especially beneficial for weaker open-source encoders.

\vspace{-1mm}
\subsection{Qualitative Answer Analysis}
\label{sec:qualitative} 
\begin{figure*}[htbp]
\centering
\includegraphics[width=0.97\linewidth]{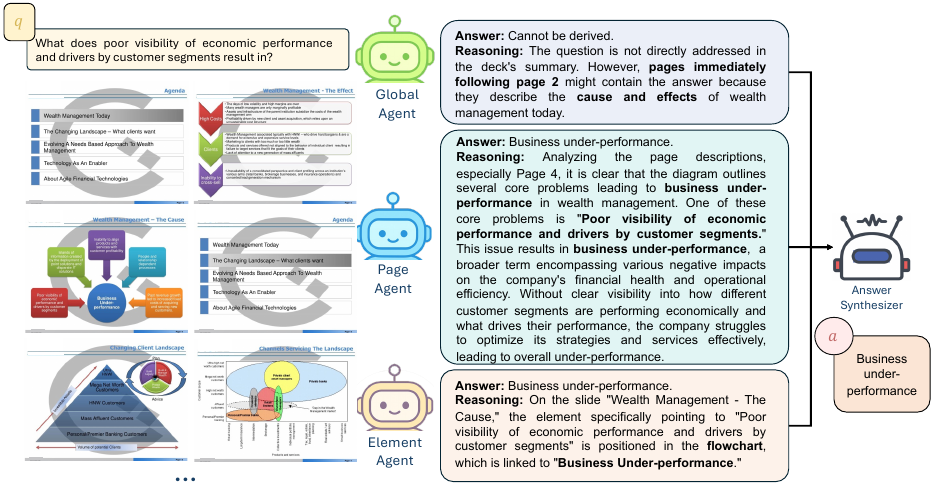}
\vspace{-2mm}
\caption{Example answers by \method. Agents at different levels collaborate for comprehensive responses.}
\label{fig:sample_answer}
\vspace{-3mm}
\end{figure*}

Figure~\ref{fig:sample_answer} shows a qualitative example of answer generated by \method to illustrate how \method leverages hierarchical knowledge for reasoning. 

Given the slide deck, \method builds hierarchical knowledge $\mathcal{K}=\{\mathcal{K}_g,\mathcal{K}_p,\mathcal{K}_e\}$ (Figures~\ref{fig:sample_global_knowledge}--\ref{fig:sample_element_knowledge}). 
Because the query lacks global/page cues (e.g. ``on page 4''), the agent orchestrator activates all agents. 
The global agent performs deck-level triage and nominates the pages directly following page 2's summary as related to ``cause/effect region.'' The page agent pinpoints \emph{Page 4: Wealth Management--The Cause}, whose description states that listed problems leads to \textsf{Business Under-Performance}. \textbf{Element} ($\mathcal{K}_e$) grounds the answer by parsing the flowchart on Page 4, extracting the node whose verbatim text matches the query and following its directed edge to the target node labeled \textsf{Business Under-Performance}. The answer synthesizer fuses these signals to return \textbf{Business under-performance} with explicit provenance (Page 4, flowchart edge from the queried node). This layering is essential: without global triage, search drifts; without page focus, we may match the phrase but miss causality; without element grounding we rely on summaries and risk speculation. 
Combining all three yields robust reasoning: the global agent provides thematic scope, the page agent narrows candidates, and the element agent confirms answers with visual grounding. This layered verification produces accurate, confident responses.
 


%% file: tables/tab-overall_gpt4o.tex
\begin{table*}[htbp]
\centering
\small
\begin{tabular}{l|ccc|ccc|ccc}
\toprule
\multirow{2}{*}{\textbf{Model}} 
& \multicolumn{3}{c|}{\textbf{SlideVQA}} 
& \multicolumn{3}{c|}{\textbf{TechSlides}} 
& \multicolumn{3}{c}{\textbf{FinSlides}} \\
& Overall & Num & F1 & Overall & Num & F1 & Overall & Num & F1 \\
\midrule
\multicolumn{10}{l}{\emph{Multimodal LLMs (Type 1)}}
\\
Gemini-2.0-Flash        & 75.0 & 71.3 & 79.8 & 50.4 & 67.6 & 41.6 & 70.8 & 70.6 & 77.8 \\
Gemini-2.5-Flash     & \underline{83.8} & \underline{78.3} & \textbf{91.8} & 51.1 & 71.4 & 41.2 & 76.2 & 75.8 & \textbf{100.0} \\
Gemini-2.5-Flash-Lite   & 71.2 & 60.8 & 87.0 & 47.3 & 58.1 & 41.9 & 57.0 & 56.6 & 68.3 \\
Claude-4.1-Opus      & 78.4 & 74.3 & 82.3 & 61.0 & \underline{81.4} & 52.3 & 56.5 & 54.8 & 73.3 \\
Claude-3.5-Haiku      & 62.5 & 68.3 & 54.6 & 52.5 & 80.2 & 39.5 & 48.5 & 49.5 & 29.6 \\
\rowcolor{lightergray}
GPT-4o            & 77.0 & 72.1 & 84.0 & 63.4 & 78.3 & 53.9 & 80.0 & 80.8 & 62.1 \\
\midrule
\multicolumn{10}{l}{\emph{Multimodal RAG (Type 2) and Agentic Methods (Type 3) with GPT-4o as base models.}}\\
COLPALI           & 78.8 & 73.7 & 83.4 & 64.1 & 73.2 & 54.5 & 80.9 & 81.5 & 62.7 \\
VisRAG            & 78.2 & 73.1 & 85.4 & 64.7 & 72.6 & 54.7 & 79.2 & 81.1 & 75.8 \\
VDocRAG           & 80.0 & 75.0 & 87.8 & 67.0 & 80.5 & 57.0 & \underline{83.5} & \underline{83.8} & 64.2 \\
ViDoRAG           & 81.1 & 76.4 & 88.1 & \underline{68.7} & 78.2 & \underline{59.4} & 82.2 & 83.3 & 65.1 \\
\rowcolor{lightergray}
SlideAgent        & \textbf{84.9} & \textbf{80.4} & \underline{90.5} & \textbf{70.9} & \textbf{82.5} & \textbf{66.2} & \textbf{85.5} & \textbf{85.9} & \underline{79.6} \\
\midrule
Impr.             & \green{+7.9}  & \green{+8.3}  & \green{+6.5}  & \green{+7.5}  & \green{+4.2}  & \green{+12.3}  & \green{+5.5}  & \green{+5.0}  & \green{+17.5}  \\
\bottomrule
\end{tabular}
\vspace{-1mm}
\caption{Performance comparison of proprietary models on SlideVQA, TechSlides, and FinSlides. 
All baseline methods use GPT-4o for answer generation. Improvements over the base model GPT-4o is shown in \green{green}. 
The best and second best performance are highlighted in \textbf{bold} and \underline{underlined}. \method outperforms all baseline methods (Type 2/3) sharing the same base model, and even outperforms stronger raw LLMs (e.g. Gemini-2.5).
late}
\label{tab:performance_proprietary}
\vspace{-2mm}
\end{table*}

%% file: tables/tab-infovqa_proprietary.tex
\begin{table}[htbp]
\centering
\small
\begin{tabular}{lccc}
\toprule
\textbf{Model} & \textbf{Overall} & \textbf{Num} & \textbf{F1} \\
\midrule
\multicolumn{4}{l}{\emph{Multimodal LLMs}} \\
Gemini-2.0-Flash  & 72.3 & 66.1 & 81.3 \\
Gemini-2.5-Flash  & \textbf{86.9} & \textbf{81.1} & \textbf{95.4} \\
Gemini-2.5-Flash Lite   & 71.7 & 62.6 & 87.0 \\
Claude-4.1-Opus & 36.3 & 35.4 & 38.9 \\
Claude-3.5-Haiku    & 31.5 & 30.8 & 37.0 \\
GPT-4o            & 69.0 & 59.3 & 90.5 \\
\midrule
\multicolumn{4}{l}{\emph{Multimodal RAG and Agentic Methods}} \\ 
ViDoRAG           & 71.2 & 60.5 & 90.7 \\
\rowcolor{lightergray}
\method        & \underline{79.6} & \underline{69.9} & \underline{94.1} \\
\midrule
Impr.             & \green{+10.6} & \green{+10.5} & \green{+3.6} \\
\bottomrule
\end{tabular}
\caption{Performance comparison among proprietary models, such as Gemini, Claude, GPT-4o, ViDoRAG, and SlideAgent on InfoVQA~\cite{mathew2022infographicvqa}.}
\label{tab:infovqa_proprietary}
\vspace{-3mm}
\end{table}

%% file: tables/tab-overall_internvl.tex
\begin{table*}[htbp]
\centering
\small
\begin{tabular}{l|ccc|ccc|ccc}
\toprule
\multirow{2}{*}{\textbf{Model}} 
& \multicolumn{3}{c|}{\textbf{SlideVQA}} 
& \multicolumn{3}{c|}{\textbf{TechSlides}} 
& \multicolumn{3}{c}{\textbf{FinSlides}} \\
& Overall & Num & F1 & Overall & Num & F1 & Overall & Num & F1 \\
\midrule
\multicolumn{10}{l}{\emph{Multimodal LLMs (Type 1)}}
\\
Llama-3.2-11B-Vision-Instruct   & 42.9  & 43.3 & 42.3 & 41.4 & 52.5 & 36.2 & 23.3 & 23.2 & 26.2 \\
Phi-3-vision-128k-instruct             & 72.3  & 61.8 & 90.6 & 59.4 & 60.0 & 59.1 & 48.8 & 48.5 & 64.3 \\
Qwen2.5-VL-7B-Instruct      & \textbf{79.5}  & \underline{70.5} & \textbf{94.3} & 59.3 & 52.5 & \textbf{65.9} & 53.6 & 52.5 & \underline{85.7} \\
Qwen2.5-VL-32B-Instruct      & \underline{79.2}  & \textbf{71.1} & \underline{92.2} & \textbf{67.5} & \textbf{87.5} & 60.6 & \underline{57.4} &  \underline{56.6} & \textbf{87.5} \\
llava-1.5-7b-hf    & 36.8  & 22.4 & 79.0 & 23.3 & 12.5 & 38.7 & 10.7 & 10.1 & 16.6 \\
llava-1.5-13b-hf    & 44.9  & 25.1 & 81.8 & 28.1 & 17.5 & 45.0 & 20.6 & 16.5 & 36.7 \\
llava-v1.6-mistral-7b-hf     & 50.9  & 37.3 & 82.6 & 34.4 & 37.5 & 32.2 & 12.2 & 12.1 & 17.8 \\
llava-v1.6-vicuna-13b-hf    & 16.7  & 10.2 & 81.5 & 45.2 & 40.0 & 49.1 & 32.0 & 31.3 & 64.3 \\
\rowcolor{lightergray}
InternVL3-8B     & 63.0  & 56.5 & 74.1 & 55.4 & 57.5 & 54.4 & 49.8 & 49.5 & 64.3 \\
\midrule
\multicolumn{10}{l}{\emph{Multimodal RAG (Type 2) and Agentic Method (Type 3) with InternVL3-8B as base models.}}\\
COLPALI          & 63.4  & 56.7 & 73.8 & 57.1 & 60.9 & 55.2 & 50.4 & 49.3 & 65.7 \\
VisRAG           & 63.6  & 56.5 & 75.5 & 56.8 & 57.7 & 55.4 & 51.1 & 49.6 & 65.2 \\
VDocRAG          & 65.2  & 59.7 & 77.0 & 59.2 & 60.7 & 58.3 & 51.8 & 50.1 & 65.9 \\
ViDoRAG          & 68.8  & 61.9 & 77.3 & 61.4 & 61.9 & 59.3 & 52.7 & 55.4 & 66.6 \\
\rowcolor{lightergray} 
SlideAgent       & 72.7  & 68.2 & 79.4 & \underline{63.1} & \underline{78.0} & \underline{61.7} & \textbf{63.3} & \textbf{62.8} & 68.3 \\
\midrule
Impr.            & \green{+9.8}  & \green{+11.7}  & \green{+5.4}  & \green{+7.7}  & \green{+20.5}  & \green{+2.3}  & \green{+13.5}  & \green{+13.3}  & \green{+4.0}  \\
\bottomrule
\end{tabular}
\vspace{-1mm}
\caption{Performance (correctness \%) of various models on SlideVQA, TechSlides, and FinSlides dataset. \method outperforms all baseline methods (Type 2/3) sharing the same LLM backbone.}
\label{tab:performance_open_source}
\vspace{-4mm}
\end{table*}

%% file: src/related.tex
\vspace{-1mm}
\section{Related Work}
\vspace{-1mm}
In this section, we discuss relevant works in visual document understanding and Multimodal LLMs. A more detailed description is in Appendix~\ref{app:related_extended}. 

\noindent \textbf{Visual Document Understanding} 
relied on computer vision pipelines combining OCR, layout parsing, and heuristics to extract semantics~\cite{shilman2005learning,bhowmik2023document,gao2019icdar,kieninger1998table,smith2007overview}. Though effective, these methods were brittle to noisy and visually rich inputs. 
Recent models such as BERTgrid~\cite{denk2019bertgrid},  LayoutLM~\cite{xu2020layoutlm,xu2020layoutlmv2}, LayoutT5~\cite{tanaka2021visualmrc}, and TILT~\cite{powalski2021going} jointly encode texts, layouts, and visuals through multimodal pretraining--laying the foundation for LLM document understanding.
\noindent \textbf{Visual Question Answering based on Multi-page Document} requires layout comprehension and long-context reasoning. Earlier studies focused on segmentation~\cite{haurilet2019spase} and generation~\cite{sun2021d2s,fu2022doc2ppt} perspectives, while recent work focus on multi-hop, numerical, or commonsense reasoning capabilities~\cite{tanaka2023slidevqa, mathew2022infographicvqa,ma2024mmlongbench}. Retrieval pipelines such as ColPali~\cite{faysse2024colpali}, VisRAG~\cite{yu2024visrag}, and VDocRAG~\cite{tanaka2025vdocrag} combine retrieval and reasoning for multi-page comprehension, emphasizing the need for fine-grained element-level reasoning.

\noindent \textbf{General-Purpose Multimodal LLMs} such as GPT-4/4o~\cite{achiam2023gpt,hurst2024gpt}, Gemini~\cite{team2023gemini}, LLaVA~\cite{liu2023visual}, and InternVL3~\cite{zhu2025internvl3exploringadvancedtraining} have advanced visual reasoning~\cite{shao2024visual,yang2023set}. 
Yet, trained largely on natural images~\cite{wu2024next}, these models struggle with visual documents such as slides, which require precise grounding of heterogeneous elements (charts, tables, icons). 

%% file: src/conclusion.tex
\vspace{-1mm}
\section{Conclusion \& Future Work}
\vspace{-1mm}
\label{sec:conclusion}
We present \method, a hierarchical agentic framework that leverages specialized agents at multiple granularity levels for multi-page visual document understanding, particularly for slide decks.
Future work can improve parsing robustness on low-contrast layouts, model inter-element relations via graphs, replace the first-$N$-pages heuristic with content-aware page selection for $\mathcal{K}_g$, and extend \method to multi-turn scenarios. 

%% file: src/limitation.tex
\section{Limitations} 
\emph{1) Element-level Retrieval.} 
We assess element-level grounding on a manually annotated subset of SlideVQA questions, providing direct evidence of fine-grained retrieval. However, element boundaries may vary across OCR and layout-parsing tools.  Broader benchmarks with element-level annotations would enable more comprehensive evaluation. \emph{2) Retrieval Method.}
Our framework supports both textual and multimodal retrieval. For efficiency, we primarily adopt text-based retrieval. Future work could explore multimodal retrievers or domain-specific retrieval strategies.

%% file: src/disclaimer.tex
\section*{Disclaimer}
\vspace{-1mm}
This paper was prepared for informational purposes by the Artificial Intelligence Research group of JPMorgan Chase \& Co and its affiliates ("J.P. Morgan") and is not a product of the Research Department of J.P. Morgan.  J.P. Morgan makes no representation and warranty whatsoever and disclaims all liability, for the completeness, accuracy or reliability of the information contained herein.  This document is not intended as investment research or investment advice, or a recommendation, offer or solicitation for the purchase or sale of any security, financial instrument, financial product or service, or to be used in any way for evaluating the merits of participating in any transaction, and shall not constitute a solicitation under any jurisdiction or to any person, if such solicitation under such jurisdiction or to such person would be unlawful. 

\section*{Acknowledgements}
The authors would like to thank Nishan Srishankar, Kelly Patel, and Manuela Veloso at J.P. Morgan AI Research for their valuable discussions and feedback on this work.

%% file: src/appendix.tex
\input{src/related_long}

\section{Additional Method Details}

\subsection{Element Merging}
\label{app:merge}
Raw output, especially text spans from the element processing pipeline (Section~\ref{sec:knowledge}) can fragment coherent texts into multiple parts, creating challenges for downstream understanding. We thus adopt a graph-based merging algorithm to reconstruct semantically coherent text blocks while preserving spatial layout.

\paragraph{Distance-Based Adjacency}
Our merging criterion is based on minimum distance between bounding boxes. For two boxes $b_i = (x_1^i, y_1^i, x_2^i, y_2^i)$ and $b_j = (x_1^j, y_1^j, x_2^j, y_2^j)$, the minimum distance is computed from the horizontal and vertical distances $d_h$ and $d_v$:

$$d_{\text{min}}(b_i, b_j) = \sqrt{d_h^2 + d_v^2}$$

where $d_h = \min(|x_2^i - x_1^j|, |x_2^j - x_1^i|)$ if boxes don't overlap horizontally, and $d_h=0$ otherwise.
Similarly, $d_v = \min(|y_2^i - y_1^j|, |y_2^j - y_1^i|)$ if boxes don't overlap vertically, and $d_v = 0$ otherwise.

Two boxes are considered adjacent if $d_{\text{min}}(b_i, b_j) \leq \tau$ where $\tau$ is a threshold we choose to be 15 pixels. The results are fed into the following graph-based component detection.

\noindent \textbf{Adjacency Graph Construction.} An undirected graph $G = (V, E)$ is constructed, where vertices $V = \{1, 2, \ldots, |B|\}$ represent the bounding boxes for the textual elements, and $E$ edges connect adjacent boxes. An edge $(i,j) \in E$ exists if $d_{\text{min}}(b_i, b_j) \leq \tau$.

\noindent \textbf{Connected Component Detection.} We identify coherent text spans using depth-first search (DFS) to find connected components. Each component $C_k = \{i_1, i_2, \ldots, i_m\}$ represents indices of boxes that should be merged into a single text block.

\noindent \textbf{Spatial Merging and Text Concatenation.} For each component $C_k$, we compute the unified bounding box:
$$B_k = \left(\min_{i \in C_k} x_1^i, \min_{i \in C_k} y_1^i, \max_{i \in C_k} x_2^i, \max_{i \in C_k} y_2^i\right)$$

The final text is given by concatenating text in the reading order. This is achieved by sorting boxes within each component using lexicographic ordering: primary sort by vertical position ($y_1$), secondary sort by horizontal position ($x_1$). 

This approach effectively reduces element fragmentation while preserving spatial relationships, enabling more accurate element-level descriptions for our hierarchical framework. Algorithm~\ref{alg:merge_boxes} (at the end of the appendix) provides the complete implementation details.

\subsection{Answer Matching}
\label{app:answer_matching}
To determine answer equivalence, i.e., if two agents agree upon their opinions, we do fuzzy string matching between each pair of answers $a_1, a_2$ using Normalized Levenshtein Similarity (NLS):
\begin{equation}
\text{NLS}(a_1, a_2) = \left(1 - \frac{\text{editdistance}(a_1, a_2)}{\max(|a_1|, |a_2|)}\right),
\end{equation}

where $\text{editdistance}(a_1, a_2)$ computes the Levenshtein distance~\citep{levenshtein1966binary} between the answers after tokenization. Two answers are considered equivalent if $\text{NLS}(a_1, a_2) \geq 0.75$.

\section{Extended Experimental Setup}
\subsection{Dataset Descriptions}
\label{app:dataset}

\begin{itemize}
    \item SlideVQA~\citep{tanaka2023slidevqa} is a multi-modal, multi-image VQA dataset consisting of $\ge$ 2600 slide decks that require single-hop, multi-hop, and numerical reasoning. Each slide deck corresponds to multiple questions. The license is available at \footnote{\url{https://github.com/nttmdlab-nlp/SlideVQA/blob/main/LICENSE}}.
    \item InfoVQA~\citep{mathew2022infographicvqa} is a dataset that features questions that require reasoning and arithmetic skills. It contains over 30,000 questions with over 5,400 images. The dataset is released under CC-BY license.
    \item TechSlides and FinSlides~\citep{wasserman2025real} are from the REAL-MM-RAG benchmark, which comprises slides and documents with text, figures, tables, and images, requiring systems to handle combined textual and visual data. The dataset is released under the CDLA-Permissive-2.0 license. 
\end{itemize}

\subsection{Metrics}
\label{app:metrics}
For ranking evaluation, we adopt MRR, Hit@k, and nDCG@k. Mean Reciprocal Rank (MRR) reflects how early the correct answer appears in the ranked list by averaging the reciprocal of its first relevant position across all queries. Hit@k measures the proportion of queries for which at least one correct answer occurs within the top-$k$ retrieved results. Normalized Discounted Cumulative Gain (nDCG@k) assesses overall ranking quality by weighting relevant items according to their positions and normalizing by the ideal ranking, thereby capturing both accuracy and ordering of retrieved answers.

\subsection{Implementation Details}
\label{app:implementation_details}
We used \texttt{EasyOCR}\footnote{\url{https://github.com/JaidedAI/EasyOCR}} and \texttt{Docling}~\citep{auer2024docling} for detecting textual and visual elements, respectively. Whenever applicable, the answers are generated using a temperature of 0.0 to ensure deterministic results.

\section{Additional Experiments}

\subsection{InfoVQA with Open-Source Backbones}
Table~\ref{tab:infovqa_open_source} reports InfoVQA performance when \method is paired with open-source MLLMs, complementing the proprietary-model results in the main paper.

\input{tables/tab-infovqa_open-source}

\subsection{Generalizability Across Models}
\method's framework is model-agnostic and provides consistent improvements when applied to stronger base models. Table~\ref{tab:gemini-qwen} shows results with recent state-of-the-art models. With Gemini-2.5-Flash, \method achieves 90.5\% overall accuracy (+6.7 points), 87.6\% numeric (+9.3 points), and 93.6\% F1 (+1.8 points) on SlideVQA, outperforming all models in Tables~\ref{tab:performance_proprietary}-\ref{tab:performance_open_source}. With Qwen2.5-VL-7B (released Feb 2025), \method achieves 80.4\% overall (+0.9 points), 73.8\% numeric (+3.3 points), and 94.5\% F1 (+0.2 points).

\input{tables/tab-gemini-qwen}

\subsection{Generalization to Other Document Types}
To validate that \method's hierarchical approach generalizes beyond slide decks, we evaluated on DocVQA~\citep{mathew2021docvqa}, a widely-used benchmark for document understanding with scanned documents, forms, and reports (Table~\ref{tab:docvqa}). With GPT-4o as base model, \method achieves 94.5\% overall accuracy (+0.9 points), 91.6\% numeric (+2.5 points), and 97.5\% F1 (+0.3 points) compared to the baseline. With InternVL3-8B, \method achieves 94.7\% overall (+0.5 points), 93.4\% numeric (+0.9 points), and 96.5\% F1 (+0.5 points).

\input{tables/tab-docvqa}

\subsection{Computational Cost and Scalability}
\label{sec:cost}
An important aspect of \method's design is that computational cost largely depends on the retriever used, which can be swapped based on latency requirements. Table~\ref{tab:cost} compares different retriever configurations versus single-shot GPT-4o.

\paragraph{Knowledge Construction (One-time, Amortized)}
OCR/layout parsing takes $2.54 \pm 0.32$ seconds per page (parallelized). Generating hierarchical knowledge ($\mathcal{K}_g, \mathcal{K}_p, \mathcal{K}_e$) takes 51.5 seconds total for a typical slide deck, using $1544 \pm 36$ MB memory. This is a one-time cost per document; the generated knowledge acts as a ``cache'' reusable across multiple queries.

\paragraph{Retrieval \& QA (Per-Query)}
\method+BM25 achieves similar latency to raw GPT-4o (9.44s vs 9.12s) but with a 91\% reduction in input tokens (1,330 vs 15,420 tokens), translating to substantial cost savings and scalability to longer documents. \method+SFR provides higher retrieval quality at 2.1$\times$ slower latency (19.17s) but still maintains 90\% token reduction. For multi-query scenarios, \method becomes increasingly cost-effective as the one-time knowledge construction is amortized across queries.

\section{Use of AI Assistants}
We used AI assistants to improve the paper’s presentation, including language polishing, clarity, and organization. AI assistants also provided support for data cleaning and processing. All technical experiments, analyses, and final content were verified by humans.

\input{tables/tab-cost}


\input{src/ethics}


\input{tables/tab-models}

\input{tables/tab-notation}

\input{tables/tab-gt_pages_gpt4o}

\input{tables/tab-gt_pages_internvl}

\input{tables/tab-global-refine}

\input{tables/tab-global-desc}

\input{tables/tab-element-desc}

\input{algorithms/alg-merge_bboxes}

\begin{figure*}[htbp]
\centering
\includegraphics[width=0.97\linewidth]{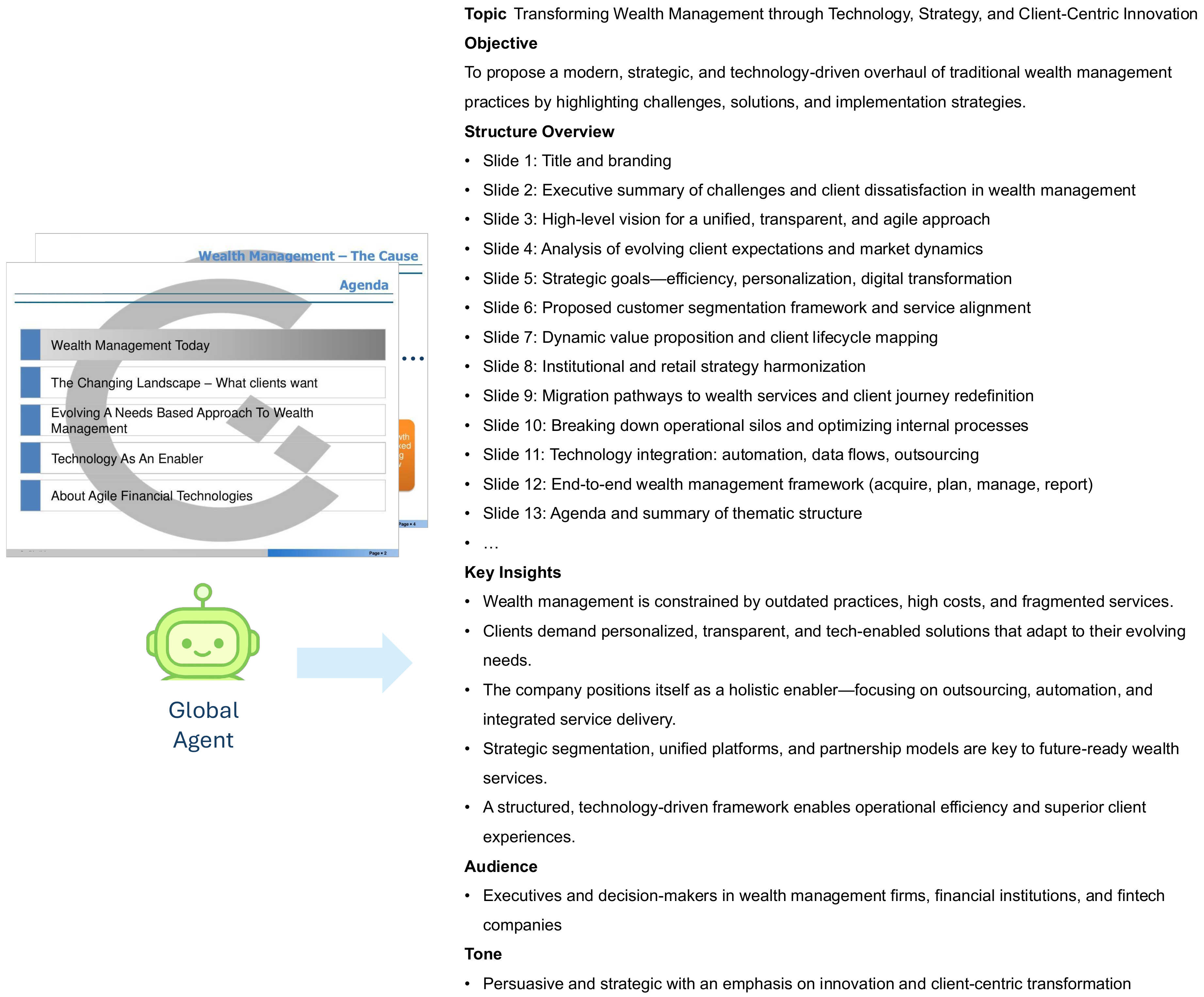}
\caption{Sample global knowledge $\mathcal{K}_g$ generated by the global agent, showing document-level summary, objectives, and narrative structure.}
\label{fig:sample_global_knowledge}
\vspace{-3mm}
\end{figure*}

\begin{figure*}[htbp]
\centering
\includegraphics[width=0.97\linewidth]{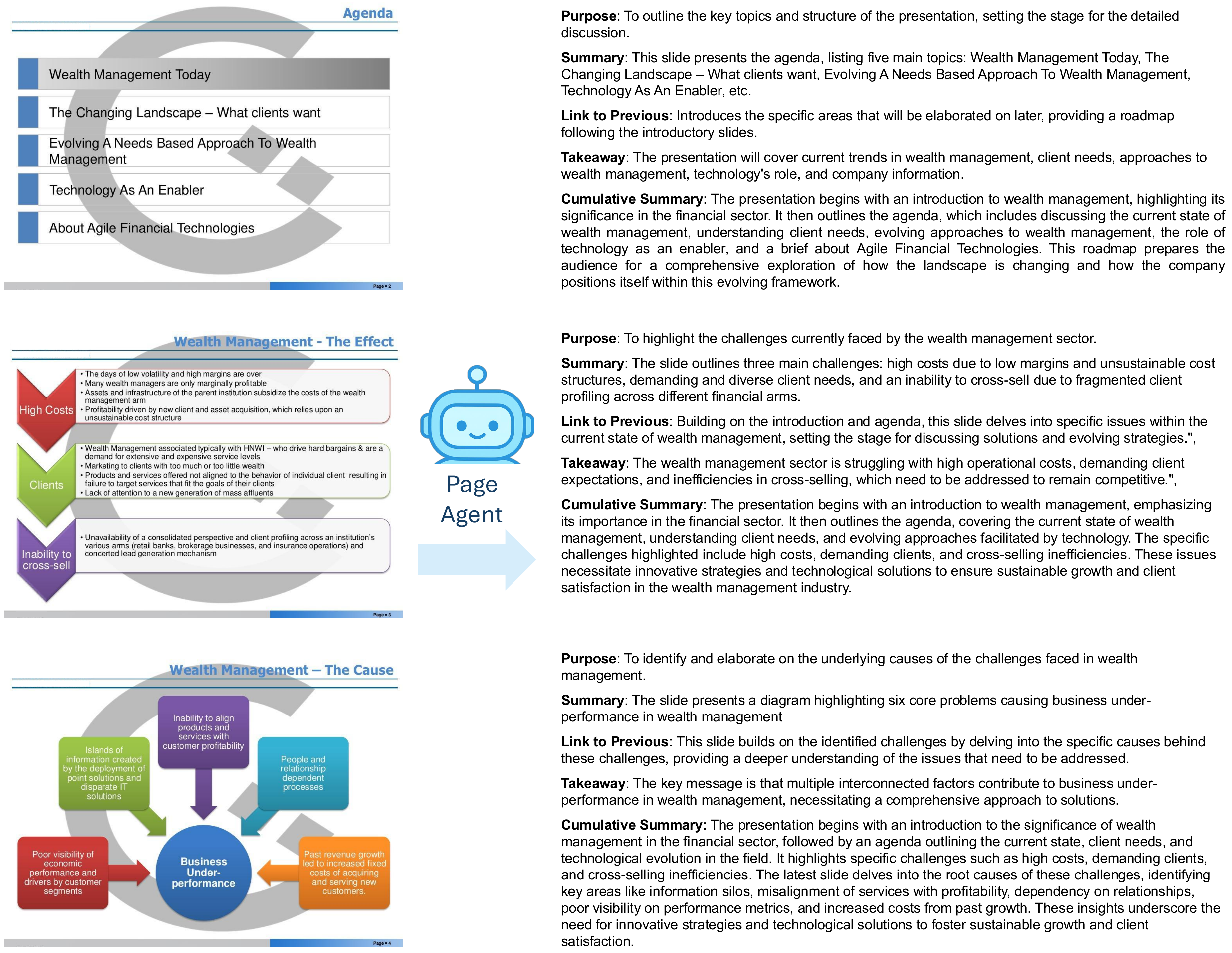}
\caption{Sample page-level knowledge representation $\mathcal{K}_p^{i}$ generated by the page agent, capturing slide-specific content and cross-slide relationships.}
\label{fig:sample_page_knowledge}
\vspace{-3mm}
\end{figure*}

\begin{figure*}[htbp]
\centering
\includegraphics[width=0.97\linewidth]{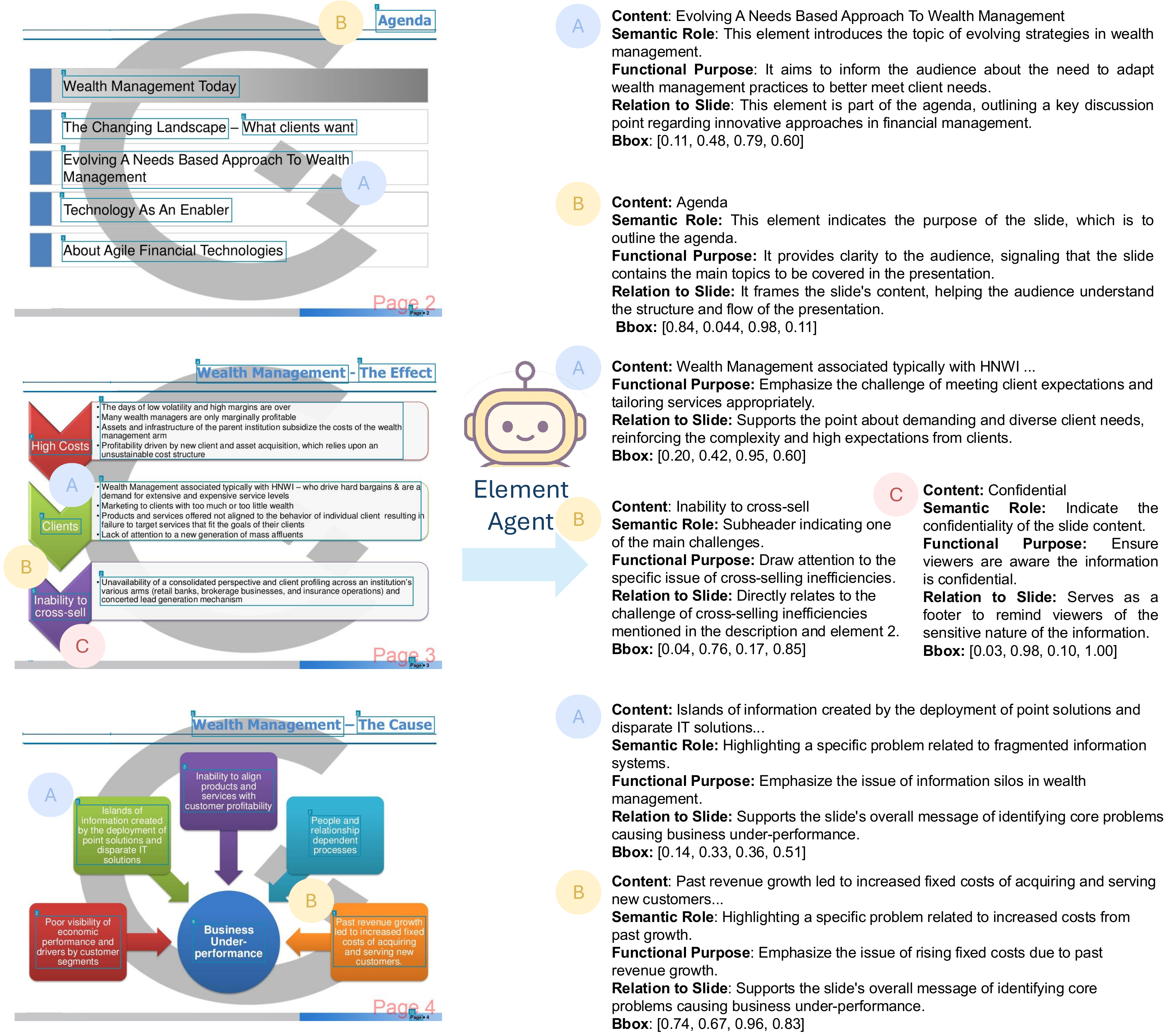}
\caption{Sample element-level knowledge representation $\mathcal{K}_e^{j}$ generated by the element agent.}
\label{fig:sample_element_knowledge}
\vspace{-3mm}
\end{figure*}

%% file: src/related_long.tex
\section{Extended Related Work}
\label{app:related_extended}
\subsection{Document and Infographics Understanding}
Early research in document and infographic understanding predates multimodal LLMs and was dominated by computer vision pipelines. Classical approaches typically combined OCR, layout parsing, and heuristic rules to extract semantics from text-rich visual content such as forms, tables, and scientific figures~\cite{shilman2005learning,bhowmik2023document,gao2019icdar,kieninger1998table,smith2007overview}. While these methods enabled structured information extraction, they were brittle to domain shifts and noisy scans.

With the rise of deep learning, several specialized benchmarks spurred progress in document visual question answering (VQA). Datasets such as DocVQA~\cite{mathew2021docvqa}, DocCVQA~\cite{tito2021document}, FUNSD~\cite{jaume2019funsd}, and DocBank~\cite{li2020docbank} emphasized reasoning over diverse document layouts, scanned forms, and large-scale document collections. ChartQA~\cite{masry2022chartqa} further extended this line by introducing reasoning over structured figures like bar and line charts.
On the modeling side, early work on text-centric VQA demonstrated that reading and interpreting embedded text was crucial~\cite{singh2019towards,mishra2019ocr}. To better represent visually rich documents, approaches such as BERTgrid~\cite{denk2019bertgrid} and LayoutLM~\cite{xu2020layoutlm} proposed contextualized embeddings that jointly encode textual content and 2D spatial layout. This line was extended by LayoutLMv2~\cite{xu2020layoutlmv2}, LayoutT5~\cite{tanaka2021visualmrc}, and TILT~\cite{powalski2021going}, which integrated stronger visual features and multimodal pretraining objectives. These advances have achieved impressive results on single-image document VQA tasks by unifying textual, layout, and visual signals, laying the foundation for more recent LLM-based systems.

\subsubsection{Slide Deck Understanding}
Building on document and infographic understanding, research has also explored presentation slides, which pose unique challenges due to their combination of dense text, figures, and multi-page structure. Early efforts addressed component-level analysis, such as object segmentation on slide pages~\cite{haurilet2019spase}, or generation tasks like creating slides from research papers~\cite{sun2021d2s,fu2022doc2ppt}. More recently, benchmarks such as MMLongBench~\cite{ma2024mmlongbench} have emphasized long-context processing over lengthy multi-page PDFs, highlighting the scalability issues that arise when moving from single documents to multi-slide collections.

\subsubsection{Slide Visual Question Answering}
Slide visual question answering (Slide VQA) emerges as a prominent direction within this broader area, aiming to automatically interpret presentation slides to answer natural-language queries. Early work such as SlideVQA~\cite{tanaka2023slidevqa} introduced multi-modal slide decks paired with questions requiring single-hop, multi-hop, and numerical reasoning across slides. This line was extended by InfoVQA~\cite{mathew2022infographicvqa}, which targeted information-centric slides and documents, often involving arithmetic and commonsense reasoning. More recent benchmarks such as TechSlides and FinSlides~\cite{wasserman2025real} from the REAL-MM-RAG suite emphasize domain-specific contexts—technical and financial presentations that integrate text, figures, and tables.

At the same time, multi-image and multi-context evaluation benchmarks such as MIBench~\cite{liu2024mibench} and MC-Bench~\cite{xu2024mc} highlight the unique reasoning challenges in multi-slide and multi-panel scenarios. Retrieval-based methods such as ColPali~\cite{faysse2024colpali} improve slide-grounded search and QA, while pipeline approaches like VisRAG~\cite{yu2024visrag} and VDocRAG~\cite{tanaka2025vdocrag} integrate retrieval with reasoning to support multi-page comprehension. Collectively, these datasets and methods underscore both the promise and complexity of slide understanding, while leaving open the question of whether element-level reasoning—beyond page-level retrieval—can substantially improve performance.

\subsubsection{General Purpose LLMs}
General-purpose large language models (MLLMs) such as GPT-4~\cite{achiam2023gpt}, GPT-4o~\cite{hurst2024gpt}, Gemini~\cite{team2023gemini}, LLaVA~\cite{liu2023visual}, InternVL3~\cite{zhu2025internvl3exploringadvancedtraining}, and Visual CoT~\cite{shao2024visual} have significantly advanced visual reasoning across a wide range of tasks. Visual prompting methods like Set-of-Marks~\cite{yang2023set} enhance reasoning by augmenting inputs with structured visual annotations. Multi-agent systems~\cite{wang2025companioncast,wang2026mascot} have also enhanced the multi-perspective reasoning capabilities of LLM-powered systems. While these open-source and closed-source MLLMs form the foundation for many comparisons, they often struggle when applied to slides, where precise grounding and fine-grained interpretation are required to parse heterogeneous elements such as charts, tables, and icons.

Yet key challenges remain. Domain-specific visual semantics are often overlooked, as LLMs trained primarily on natural images~\cite{wu2024next} struggle to capture the specialized conventions of slides and infographics--for instance, repetitive logos, color-coded encodings, or abstract symbolic icons. Metadata-dependent integration is similarly fragile: real-world slides and PDFs frequently lack clean structural annotations, and scanned or flattened exports strip away layout cues, rendering metadata-reliant systems brittle~\cite{huridocs2025pdf,rausch2021docparser}. Finally, scalable fine-grained reasoning capabilities remains limited~\cite{jin2024agentreview}. Current models can process only a small number of images at once~\cite{liu2023visual}, while retrieval typically occurs at the page level~\cite{faysse2024colpali,tanaka2025vdocrag}, overlooking element-level reasoning.

Our work builds on these developments by introducing a metadata-free, slide-specialized agentic framework. Unlike prior page-level approaches, our method constructs hierarchical representations and employs retrieval-then-reasoning at the global (deck-level), page (slide-level), and element (fine-grained component) layers. This design enables robust comprehension across diverse slide domains and opens the door to systematic evaluation of whether element-level reasoning provides measurable improvements over page-level retrieval alone.

%% file: tables/tab-infovqa_open-source.tex
\begin{table}[htbp]
\centering
\small
\begin{tabular}{lccc}
\toprule
\textbf{Model} & \textbf{Overall} & \textbf{Num} & \textbf{F1} \\
\midrule
\multicolumn{4}{l}{\emph{Raw LLMs}} \\
Llama-3.2-11B-Vision-Instruct    & 55.4 & 49.6 & 67.9 \\
Phi-3-vision-128k-instruct               & 53.4 & 43.1 & 88.7 \\
Qwen2.5-7B-Instruct         & \textbf{80.1} & \textbf{73.0} & \textbf{96.4} \\
Qwen2.5-32B-Instruct       & \underline{72.1} & \underline{62.0} & \underline{94.8} \\
llava-1.5-7b-hf       & 25.1 & 10.6 & 83.9 \\
llava-1.5-13b-hf     & 20.2 & 8.0  & 84.2 \\
llava-v1.6-mistral-7b-hf      & 32.6 & 22.6 & 79.9 \\
llava-v1.6-vicuna-13b-hf      & 34.9 & 24.8 & 80.9 \\
\rowcolor{lightergray}
InternVL3-8B       & 66.7 & 57.7 & 85.2 \\
\midrule
\multicolumn{4}{l}{\emph{Multimodal RAG and Agentic Methods}} \\ 
ViDoRAG            & 67.1 & 59.2 & 85.9 \\
\rowcolor{lightergray}
SlideAgent         & \underline{75.4} & \underline{66.5} & \underline{92.1} \\
\midrule
Impr.              & \green{+8.7} & \green{+8.8} & \green{+6.8} \\
\bottomrule
\end{tabular}
\vspace{-1mm}
\caption{Performance comparison among open-source multimodal models and agentic methods.}
\label{tab:infovqa_open_source}
\vspace{-3mm}
\end{table}

%% file: tables/tab-gemini-qwen.tex
\begin{table}[t]
\centering
\small
\begin{tabular}{lccc}
\toprule
\textbf{Model} & \textbf{Overall} & \textbf{Num} & \textbf{F1} \\
\midrule
\multicolumn{4}{l}{\textit{Gemini-2.5-Flash on SlideVQA}} \\
Raw Model & 83.8 & 78.3 & 91.8 \\
+ SlideAgent & \textbf{90.5} & \textbf{87.6} & \textbf{93.6} \\
Improvement & +6.7 & +9.3 & +1.8 \\
\midrule
\multicolumn{4}{l}{\textit{Qwen2.5-VL-7B-Instruct on SlideVQA}} \\
Raw Model & 79.5 & 70.5 & 94.3 \\
+ SlideAgent & \textbf{80.4} & \textbf{73.8} & \textbf{94.5} \\
Improvement & +0.9 & +3.3 & +0.2 \\
\bottomrule
\end{tabular}
\caption{SlideAgent with recent state-of-the-art models. Gemini-2.5-Flash + SlideAgent achieves best overall performance (90.5\%), establishing state-of-the-art on SlideVQA.}
\label{tab:gemini-qwen}
\end{table}

%% file: tables/tab-docvqa.tex
\begin{table}[t]
\centering
\small
\begin{tabular}{lccc}
\toprule
\textbf{Model} & \textbf{Overall} & \textbf{Num} & \textbf{F1} \\
\midrule
\multicolumn{4}{l}{\textit{GPT-4o as base model}} \\
GPT-4o & 93.6 & 89.1 & 97.2 \\
SlideAgent & \textbf{94.5} & \textbf{91.6} & \textbf{97.5} \\
Improvement & +0.9 & +2.5 & +0.3 \\
\midrule
\multicolumn{4}{l}{\textit{InternVL3-8B as base model}} \\
InternVL3-8B & 94.2 & 92.5 & 96.0 \\
SlideAgent & \textbf{94.7} & \textbf{93.4} & \textbf{96.5} \\
Improvement & +0.5 & +0.9 & +0.5 \\
\bottomrule
\end{tabular}
\caption{Performance on DocVQA, demonstrating generalization beyond slide decks to scanned documents, forms, and reports.}
\label{tab:docvqa}
\end{table}

%% file: tables/tab-cost.tex
\begin{table*}[t]
\centering
\small
\begin{tabular}{lcccc}
\toprule
\textbf{Model} & \textbf{Latency (s)} & \textbf{Tokens/sec} & \textbf{Memory (MB)} & \textbf{Input Tokens} \\
\midrule
Raw Model & 9.12 $\pm$ 1.12 & 0.76 $\pm$ 0.27 & 1253 $\pm$ 35 & 15420 $\pm$ 2 \\
SlideAgent (BM25) & 9.44 $\pm$ 4.51 & 0.73 $\pm$ 0.61 & 1317 $\pm$ 32 & 1330 $\pm$ 212 \\
SlideAgent (SFR) & 19.17 $\pm$ 13.74 & 0.13 $\pm$ 0.05 & 2495 $\pm$ 7 & 1554 $\pm$ 342 \\
\bottomrule
\end{tabular}
\caption{Computational cost comparison on SlideVQA. SlideAgent achieves >90\% token reduction (15420 $\rightarrow$ 1330 tokens) while maintaining competitive latency. The one-time knowledge construction (51.5s total, 1544 MB) is amortized across multiple queries.}
\label{tab:cost}
\end{table*}

%% file: src/ethics.tex
\section{Ethical Considerations}

\paragraph{Data Privacy, Consent, and Intellectual Property.} Visual documents such as those processed by \method may contain sensitive business or personal data. Ensuring compliance with privacy regulations (e.g., GDPR, CCPA) and obtaining appropriate consent are essential. Organizations should also respect intellectual property rights and establish policies that balance knowledge sharing with fair use. 

\paragraph{Content Reliability and User Responsibility.} 
Like other LLM-based systems, \method may inherit biases or produce inaccurate content. While it enhances question answering, outputs should be verified by human judgment, particularly in sensitive or high-stakes scenarios. Clear user guidance and validation practices can help ensure responsible use.

\begin{figure}
\centering
\includegraphics[width=0.97\linewidth]{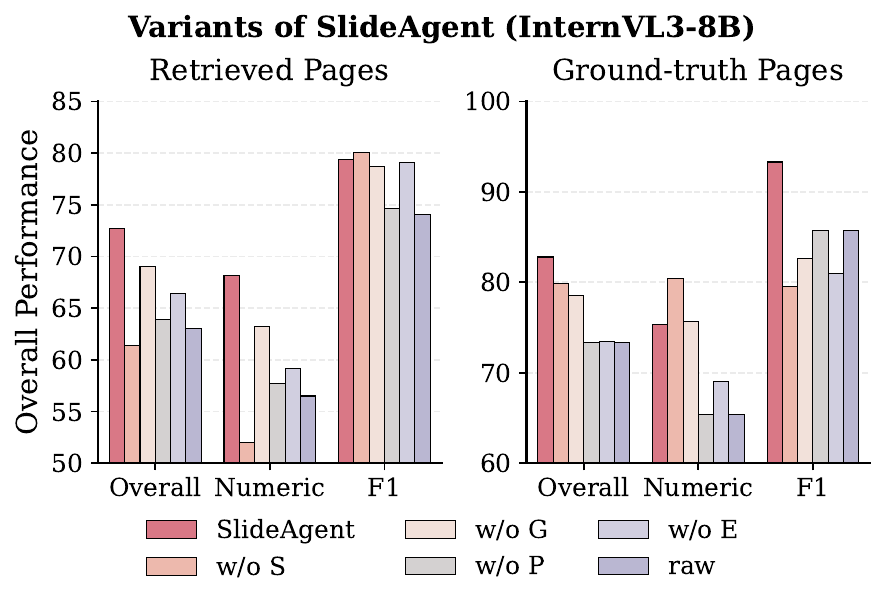}
\vspace{-2mm}
\caption{Performance comparison among variants of \method with base model InternVL3-8B. }
\label{fig:ablation_internvl3-8b}
\vspace{-2mm}
\end{figure}


%% file: tables/tab-models.tex
\begin{table*}[htbp]
\centering
\small
\renewcommand{\arraystretch}{1.4} 
\setlength{\arrayrulewidth}{0.8mm} 
\begin{tabular}{>{\raggedright\arraybackslash}p{2.5cm} p{3.5cm} p{2.5cm} p{2.5cm} p{2.5cm}}
\toprule
\textbf{Family} & \textbf{Model} & \textbf{Parameters} & \textbf{Knowledge Cutoff} & \textbf{Release} \\
\midrule
GPT~\citep{gpt4o} & gpt-4o & \textbackslash & Oct 2023 & May 2024 \\
\midrule
Gemini~\citep{team2023gemini} & gemini-2.5-flash & \textbackslash & Jan 2025 & June 2025 \\
& gemini-2.5-flash-lite & \textbackslash & Jan 2025 & June 2025 \\
& gemini-2.0-flash & \textbackslash & June 2024 & Jan 2025 \\
\midrule
Claude~\citep{claude} & claude-3-5-haiku-latest & \textbackslash & April 2024 & Oct 2024 \\ 
& claude-opus-4-1-20250805 & \textbackslash & Jan 2025 & Aug 2025 \\
\midrule
Llama~\citep{grattafiori2024llama} & Llama-3.2-11B-Vision-Instruct & 11B & \textbackslash & Sep 2024  \\
\midrule
InternVL~\citep{chen2024internvl} & InternVL3-8B & 8B & \textbackslash & Apr 2025 \\
\midrule
Phi~\citep{abdin2024phi} & Phi-3-vision-128k-instruct & 3.8B & Oct 2023 & July 2024 \\
\midrule
Qwen~\citep{bai2025qwen2} & Qwen2.5-VL-7B-Instruct & 7B & \textbackslash & Feb. 2025 \\
& Qwen2.5-VL-32B-Instruct & 32B & \textbackslash & Feb 2025 \\
\midrule
LLaVA~\citep{liu2023visual} & llava-1.5-7b-hf & 7B & Dec 2022 & Apr. 2023 \\
& llava-1.5-13b-hf & 13B & Dec 2022 & Apr. 2023 \\
& llava-v1.6-mistral-7b-hf & 7B & Dec 2023 & Mar 2024 \\
& llava-v1.6-vicuna-13b-hf & 13B & Dec 2023 & Mar 2024 \\
\bottomrule
\end{tabular}
\caption{Overview of models used in the experiments, including their family, model name, parameter size, knowledge cutoff date, and release date. }
\label{tab:models}
\end{table*}

%% file: tables/tab-notation.tex
\begin{table*}[htbp]
\centering
\begin{tabular}{l|p{0.75\linewidth}}
\toprule
\textbf{Notation} & \textbf{Description} \\
\midrule
$p_i, v_i$ & A slide page and its visual content \\
$\mathcal{P} = \{p_1, p_2, \dots\}$ & Set of pages in the slide deck \\
$q$ & User query \\
$\hat{Q}$ & Set of subqueries generated from the original query $q$ \\
$a$ & The final answer generated by \method \\
$e_j$ & Individual element in a slide, consisting of visual content, bounding box, and type \\
$b_j$ & Bounding box coordinates of an element within a slide \\
$t_j$ & Type of the element (e.g., text, chart, image, table) \\
$\mathcal{K}_g, \mathcal{K}_p, \mathcal{K}_e$ & Global, page, and element knowledge \\
$\mathcal{M}_g, \mathcal{M}_p, \mathcal{M}_e$ & Global, page, and element agent \\
$\mathcal{R}$ & Retrieval function that fetches relevant pages or elements based on subqueries \\
$\phi(\cdot)$ & Answer synthesizer that combines reasoning from all agents to generate the final answer $a$ \\
$\mathrm{Annot}(\hat{V})$ & Annotated visual content used by the element agent for processing \\
$h_g, h_p, h_e$ & Answers and reasoning from global, page, and element agents \\
\bottomrule
\end{tabular}
\caption{Mathematical notation used throughout the paper.}
\label{tab:notation}
\end{table*}

%% file: tables/tab-gt_pages_gpt4o.tex
\begin{table*}[htbp]
\centering
\begin{tabular}{l|ccc|ccc|ccc}
\toprule
\multirow{2}{*}{\textbf{Model}}
& \multicolumn{3}{c|}{\textbf{SlideVQA}}
& \multicolumn{3}{c|}{\textbf{TechSlides}}
& \multicolumn{3}{c}{\textbf{FinSlides}} \\
& Overall & Num & F1 & Overall & Num & F1 & Overall & Num & F1 \\
\midrule
\multicolumn{10}{l}{\emph{Raw Models}} \\
Gemini 2.0 & 86.3 & 81.0 & 90.3 & 59.7 & 60.0 & \underline{59.6} & 78.1 & 77.8 & \textbf{88.9}  \\ 
Gemini 2.5 & \textbf{89.0} & \textbf{85.7} & \textbf{93.2} & 61.5 & 65.0 & 59.5 & 76.6 & 76.4 & 83.3 \\ 
Gemini 2.5-lite & 81.8 & 75.5 & 90.1 & 56.3 & 55.0 & 57.0 & 78.1 & 78.4 & 66.7 \\ 
Claude 4.1 & 85.7 & 82.4 & 89.7 & 58.0 & 77.5 & 48.3 & 52.6 & 52.0 & 60.8 \\ 
Claude 3.5 & 58.2 & 64.0 & 50.9 & 55.8 & \textbf{83.7} & 42.5 & 47.8 & 48.0 & 40.3 \\ 
\rowcolor{lightergray}
GPT-4o & 79.4 & 71.9 & 86.4 & 64.5 & 77.1 & 58.0 & 83.0 & 84.3 & 80.1 \\ 
\midrule 
\multicolumn{10}{l}{\emph{Multimodal RAG and Agentic Methods}} \\ 
ViDoRAG & 81.8 & 73.8 & 87.0 & \underline{65.8} & 78.0 & 58.7 & \underline{84.2} & \underline{84.7} & 81.5 \\ 
\rowcolor{lightergray}
SlideAgent & \underline{87.1} & \underline{84.4} & \underline{90.6} & \textbf{68.7} & \underline{82.5} & \textbf{61.5} & \textbf{85.8} & \textbf{85.9} & \underline{85.6} \\ 
\midrule 
Impr. & \green{+7.7} & \green{+12.5} & \green{+4.2} & \green{+4.1} & \green{+5.4} & \green{+3.5} & \green{+2.8} & \green{+1.5} & \green{+5.5} \\ 
\bottomrule 
\end{tabular} 
\caption{Performance comparison of baselines and proprietary models on SlideVQA, TechSlides, and FinSlides, assuming access to ground-truth pages containing the correct answer. All baseline methods use GPT-4o for question-answering. } 
\label{tab:gt_pages_proprietary} 
\end{table*}

%% file: tables/tab-gt_pages_internvl.tex
\begin{table*}[htbp]
\centering
\small
\begin{tabular}{l|ccc|ccc|ccc}
\toprule
\multirow{2}{*}{\textbf{Model}}
& \multicolumn{3}{c|}{\textbf{SlideVQA}}
& \multicolumn{3}{c|}{\textbf{TechSlides}}
& \multicolumn{3}{c}{\textbf{FinSlides}} \\
& Overall & Num & F1 & Overall & Num & F1 & Overall & Num & F1 \\
\midrule
\multicolumn{10}{l}{\emph{Raw Models}} \\
Llama-3.2-11B-Vision-Instruct & 44.6 & 52.1 & 34.6 & 47.1 & 62.8 & 39.3 & 39.1 & 39.2 & 33.5 \\
Phi-3-vision-128k-instruct & 78.3 & 69.1 & 91.6 & 53.9 & 67.4 & 47.2 & \underline{63.8} & \underline{63.7} & 65.1 \\
Qwen2.5-VL-7B-Instruct & \underline{85.1} & \underline{77.7} & \textbf{95.3} & 57.7 & 69.8 & 51.8 & 52.7 & 52.0 & \textbf{77.8} \\
Qwen2.5-VL-32B-Instruct & \textbf{87.4} & \textbf{82.6} & \underline{93.7} & 49.6 & 62.8 & 43.2 & \textbf{69.5} & \textbf{69.6} & 65.1 \\
llava-1.5-7b-hf   & 42.9 & 27.9 & 79.9 & 24.6 & 15.0 & 39.4 & 14.2 & 14.4 & 20.9 \\
llava-1.5-13b-hf   & 46.7 & 29.5 & 83.1 & 29.2 & 17.5 & 46.6 & 23.8 & 20.3 & 41.2 \\
llava-v1.6-mistral-7b-hf  & 59.0 & 45.8 & 84.2 & 36.2 & 40.0 & 34.0 & 16.0 & 15.2 & 21.3 \\
llava-v1.6-vicuna-13b-hf& 62.3 & 48.2 & 87.1 & 54.7 & 62.5 & 49.5 & 35.2 & 30.7 & 67.6 \\
\rowcolor{lightergray}
InternVL3-8B & 73.3 & 65.4 & 85.7 & 58.4 & 72.1 & 51.5 & 56.3 & 55.9 & 65.6 \\
\midrule
\multicolumn{10}{l}{\emph{Baseline methods based on InternVL3-8B}} \\
ViDoRAG & 76.3 & 68.1 & 89.9 & \underline{61.3} & \underline{75.1} & \underline{53.9} & 58.4 & 58.7 & 67.3 \\
\rowcolor{lightergray}
SlideAgent & 82.8 & 75.3 & 93.3 & \textbf{64.6} & \textbf{79.5} & \textbf{58.0} & 62.8 & 62.6 & \underline{68.1} \\
\midrule
Impr. & \green{+9.5} & \green{+9.8} & \green{+7.6} & \green{+6.2} & \green{+7.4} & \green{+6.4} & \green{+6.5} & \green{+6.7} & \green{+2.5} \\
\bottomrule
\end{tabular}
\caption{Performance comparison of baselines and proprietary models on SlideVQA, TechSlides, and FinSlides, assuming access to ground-truth pages containing the correct answer. All baseline methods use InternVL3-8B for question-answering. }
\label{tab:gt_pages_internvl}
\end{table*}

%% file: tables/tab-global-refine.tex
\begin{table*}[t]
\centering
\small
\begin{tabularx}{0.97\linewidth}{p{0.18\linewidth}X}
\toprule
\textbf{Input} & Initial global knowledge $\mathcal{K}_g^{(0)}$ and concatenated page-level knowledge $\mathrm{concat}(\mathcal{K}_p^1,\ldots,\mathcal{K}_p^{|\mathcal{P}|})$. \\
\midrule
\textbf{Instruction} & Rewrite the global knowledge from the full page-level evidence. Use $\mathcal{K}_g^{(0)}$ only as a weak prior for terminology and early context. Incorporate complete document coverage and narrative flow, including conclusions, summaries, and next steps that appear in later pages. \\
\midrule
\textbf{Update Rule} & Regenerate each field rather than appending snippets. If later-page evidence contradicts or refines an earlier hypothesis, overwrite the earlier field with the better-supported document-level interpretation. \\
\midrule
\textbf{Output Fields} & Title, objective, structure overview, key insights, audience, and tone. \\
\bottomrule
\end{tabularx}
\caption{Prompt template for the global refinement operator. The operator performs a single-pass fieldwise rewrite from full-document page evidence.}
\label{tab:global_refine_prompt}
\end{table*}

%% file: tables/tab-global-desc.tex
\begin{table*}[htbp]
\begin{tabular}{p{0.95\linewidth}}
\toprule
You are given a complete slide deck consisting of multiple slides. Your task is to synthesize a concise, high-level summary of the overall message, structure, and purpose of the deck.
\\ \\
\textbf{\#\#\# Cumulative Summary of the Preceding Slides}

\{\textsf{Cumulative Summary}\} \\ \\
\textbf{\#\#\# Response format}
Please respond with a markdown string that follows the following format:

\textbf{Title}
$<$Concise Explicit / Inferred Title of the Presentation$>$ \\ \\
\textbf{Objective}   $<$What is the presentation trying to achieve? (e.g., inform, persuade, pitch, propose)$>$ \\ \\
\textbf{Structure Overview}   
\begin{itemize}
  \item \textbf{Slide 1}: $<$Brief description of the slide$>$
  \item \textbf{Slide 2}: $<$Brief description of the slide$>$
  \item \textbf{...}
  \item \textbf{Slide N}: $<$Brief description of the slide$>$
\end{itemize} 
\textbf{Key Insights}   
\begin{itemize}
  \item $<$Major takeaway 1$>$
  \item $<$Major takeaway 2$>$
  \item ...
\end{itemize} 
\textbf{Audience} $<$Intended audience type (e.g., executives, investors, engineers)$>$ \\
\textbf{Tone}   $<$Overall tone: e.g., persuasive, analytical, optimistic, urgent$>$ \\
\bottomrule
\end{tabular}
\caption{Prompt template for the global agent to generate comprehensive slide deck summaries, including title, objective, structure, key insights, audience, and tone.}
\label{tab:global_desc}
\end{table*}

%% file: tables/tab-element-desc.tex
\begin{table*}[htbp]
\centering
\begin{tabular}{p{0.99\linewidth}}
\toprule
\textbf{Element Type} $<$text | image | chart | table | icon | button | etc.$>$ \\
\textbf{Position on Slide} $<$e.g., top-right, centered, below title$>$ \\
\textbf{Verbatim Content} $<$if text, give the literal string; else describe the visual$>$ \\
\textbf{Semantic Role} $<$What is the element trying to do or communicate?$>$ \\
\textbf{Functional Purpose} $<$Its practical function within the slide (e.g., emphasize point, guide attention, show evidence, support action)$>$ \\
\textbf{Relation to Slide} $<$How does it connect to or support the slide's overall message?$>$ \\
\textbf{Inferred Importance} $<$how central is this element to the slide? Answer with low, medium, or high$>$ \\
\bottomrule
\end{tabular}
\caption{Element-level slide description prompt format used by the element agent to generate detailed annotations for each visual component.}
\label{tab:element_desc}
\end{table*}

%% file: algorithms/alg-merge_bboxes.tex
\begin{algorithm}[ht]
\caption{Graph-based depth-first search for merging fragmented elements into coherent semantic units while preserving spatial layout.}
\label{alg:merge_boxes}
\textbf{Input:} Set of OCR bounding boxes $B = \{b_i = (x_1^i, y_1^i, x_2^i, y_2^i, t_i)\}_{i=1}^{|B|}$, distance threshold $\tau = 15$ \\
\textbf{Output:} Merged bounding boxes $\tilde{B}$
\begin{algorithmic}[1]
    \State Initialize adjacency matrix $A \in \{0,1\}^{|B| \times |B|}$ with zeros
    \For{$i = 1$ to $|B|$}
        \For{$j = i + 1$ to $|B|$}
            \If{$d_{\text{min}}(b_i, b_j) \leq \tau$}
                \State $A[i,j] \leftarrow 1$, $A[j,i] \leftarrow 1$
            \EndIf
        \EndFor
    \EndFor
    \State Initialize visited array $\text{visited}[1:|B|] \leftarrow \text{false}$
    \State Initialize components list $\mathcal{C} \leftarrow []$
    \For{$i = 1$ to $|B|$}
        \If{$\text{visited}[i] = \text{false}$}
            \State Initialize component $C \leftarrow []$
            \State $\text{DFS}(i, A, \text{visited}, C)$ \Comment{Collect connected component}
            \State Append $C$ to $\mathcal{C}$
        \EndIf
    \EndFor
    \State Initialize merged boxes $\tilde{B} \leftarrow []$
    \For{each component $C \in \mathcal{C}$}
        \If{$|C| = 1$}
            \State Append $b_{C[0]}$ to $\tilde{B}$ \Comment{Single box, no merging}
        \Else
            \State $x_1 \leftarrow \min_{i \in C} x_1^i$, $y_1 \leftarrow \min_{i \in C} y_1^i$
            \State $x_2 \leftarrow \max_{i \in C} x_2^i$, $y_2 \leftarrow \max_{i \in C} y_2^i$
            \State Sort $C$ by $(y_1^i, x_1^i)$ to get reading order
            \State $t_{\text{merged}} \leftarrow \text{" ".join}(\{t_i\}_{i \in \text{sorted}(C)})$
            \State Append $(x_1, y_1, x_2, y_2, t_{\text{merged}})$ to $\tilde{B}$
        \EndIf
    \EndFor
    \State \Return $\tilde{B}$
\end{algorithmic}
\end{algorithm}